\documentclass[runningheads]{llncs}
\usepackage{graphicx}
\usepackage{comment}
\usepackage{amsmath,amssymb} % define this before the line numbering.
\usepackage{color}

\usepackage{url}

\usepackage[pagebackref=true,breaklinks=true,letterpaper=true,colorlinks,bookmarks=false]{hyperref}

\definecolor{note}{rgb}{0.3,0.3,1}
\definecolor{remove}{rgb}{0.7,0.1,0.1}
\definecolor{TODO}{rgb}{0.9,0.5,0.1}

\usepackage{tikz,pgfplots}

\usepackage{colortbl}
\definecolor{rowblue}{rgb}{0.9,0.95,1}
\definecolor{rowgray}{rgb}{0.9275,0.9275,0.9275}
\definecolor{rowfgray}{rgb}{0.9275,0.9275,0.9275}
\definecolor{rowbgray}{rgb}{0.975,0.975,0.975}
\definecolor{rowdgray}{rgb}{0.85,0.85,0.85}

\begin{document}
% \renewcommand\thelinenumber{\color[rgb]{0.2,0.5,0.8}\normalfont\sffamily\scriptsize\arabic{linenumber}\color[rgb]{0,0,0}}
% \renewcommand\makeLineNumber {\hss\thelinenumber\ \hspace{6mm} \rlap{\hskip\textwidth\ \hspace{6.5mm}\thelinenumber}}
% \linenumbers
\pagestyle{headings}
\mainmatter
\def\ECCVSubNumber{203}  % Insert your submission number here

\title{Learning Object Depth from Camera Motion and Video Object Segmentation} 
%\title{Learning to Estimate Depth from Camera Motion and Video Object Segmentation} 

% INITIAL SUBMISSION 
\begin{comment}
\titlerunning{ECCV-20 submission ID \ECCVSubNumber} 
\authorrunning{ECCV-20 submission ID \ECCVSubNumber} 
\author{Anonymous ECCV submission}
%\author{Brent A. Griffin and Jason J. Corso}
\institute{Paper ID \ECCVSubNumber}
\end{comment}
%******************

% CAMERA READY SUBMISSION
%\begin{comment}
\titlerunning{Learning Object Depth from Camera Motion and Segmentation}
% If the paper title is too long for the running head, you can set
% an abbreviated paper title here
%
\author{Brent A. Griffin \and Jason J. Corso}
\authorrunning{B. Griffin and J. Corso}
% First names are abbreviated in the running head.
% If there are more than two authors, 'et al.' is used.
%
\institute{University of Michigan \\%, Ann Arbor MI 48109, USA \\
\email{\{griffb,jjcorso\}@umich.edu}}
%\end{comment}
%******************
\maketitle

\begin{abstract}
Video object segmentation, i.e., the separation of a target object from background in video, has made significant progress on real and challenging videos in recent years.
To leverage this progress in 3D applications, this paper addresses the problem of learning to estimate the depth of segmented objects given some measurement of camera motion (e.g., from robot kinematics or vehicle odometry).
We achieve this by, first, introducing a diverse, extensible dataset and, second, designing a novel deep network that estimates the depth of objects using only segmentation masks and uncalibrated camera movement.
Our data-generation framework creates artificial object segmentations that are scaled for changes in distance between the camera and object, and our network learns to estimate object depth even with segmentation errors.
We demonstrate our approach across domains using a robot camera to locate objects from the YCB dataset and a vehicle camera to locate obstacles while driving.

\end{abstract}

\section{Introduction}

Perceiving environments in three dimensions (3D) is important for locating objects, identifying free space, and motion planning in robotics and autonomous vehicles.
Although these domains typically rely on 3D sensors to measure depth and identify free space (e.g., LiDAR \cite{GaEtAl20} or RGBD cameras \cite{FeLa19}), classifying and understanding raw 3D data is a challenging and ongoing area of research \cite{KhEtAl19,SuSh19,LiuEtAl19,pointnet}. %\cite{voxelnet,pointnet,modelnet,KhEtAl19,SuSh19,LiEtAl19}.
Alternatively, RGB cameras are less expensive and more ubiquitous than 3D sensors, and there are many more datasets and methods based on RGB images \cite{ImageNet,maskrcnn,MSCOCO}.
Thus, even when 3D sensors are available, RGB images remain a critical modality for understanding data and identifying objects \cite{FlCoGr19,WaEtAl19}. %boundaries and apply semantic labels to data {\color{note}(cite)}.

To identify objects in a sequence of images, video object segmentation (VOS) addresses the problem of densely labeling target objects in video.
VOS is a hotly studied area of video understanding, with frequent developments and improving performance on challenging VOS benchmark datasets \cite{SegTrackv2,DAVIS,DAVIS17,SegTrack,YTVOS}.
These algorithmic advances in VOS support learning object class models~\cite{OnReVeECCV2014,TaSuYaCVPR2013}, scene parsing~\cite{LiHeCVPR2015,TiLaIJCV2012}, action recognition~\cite{LuXuCoCVPR2015,SoIdShICCV2015,SoIdShCVPR2016}, and video editing applications~\cite{ChChChACMM2012}.

\begin{figure} [t]
	\centering
	\includegraphics[width=0.975\textwidth]{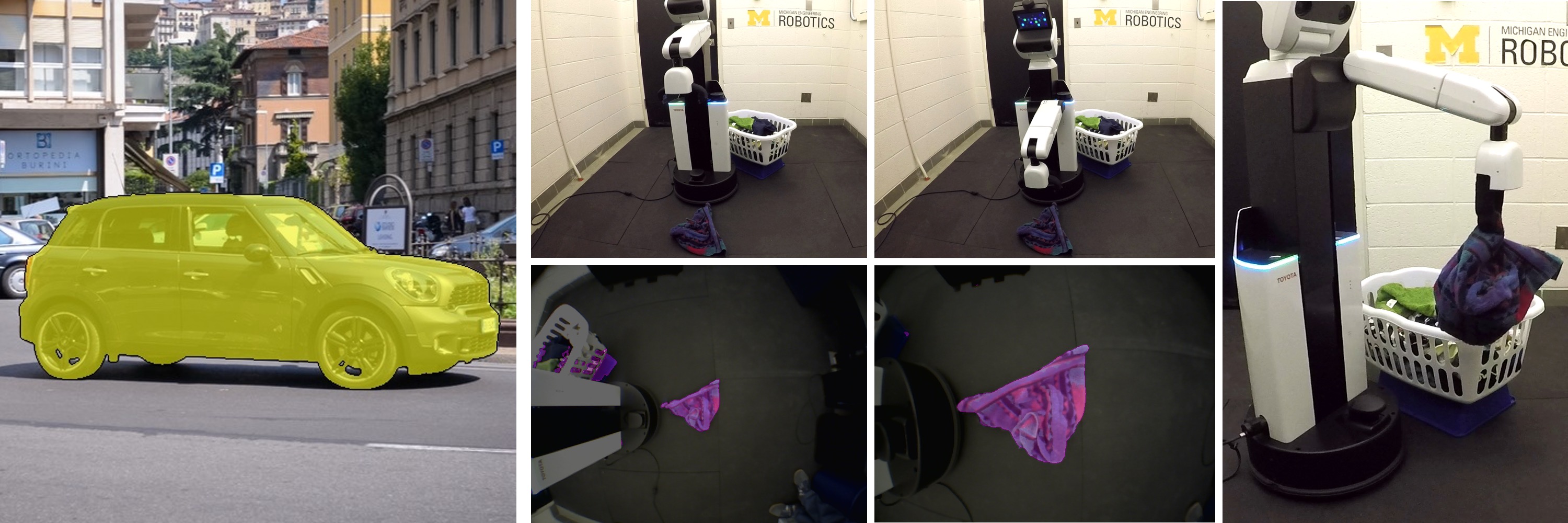}
	\caption{ \textbf{Depth from Video Object Segmentation.}
		Video object segmentation algorithms can densely segment target objects in a variety of settings (DAVIS \cite{DAVIS}, \textit{left}).
		Given object segmentations and a measure of camera movement (e.g., from vehicle odometry or robot kinematics, \textit{right}), our network can estimate an object's depth % using a single RGB camera
		%		Given a measure of camera movement (e.g., from vehicle odometry or robot kinematics, \textit{right}), our method additionally estimates an object's depth using a single RGB camera
	}
	\label{fig:front_vos}
\end{figure}

\begin{figure} [t]
	\centering
	\includegraphics[width=0.75\textwidth]{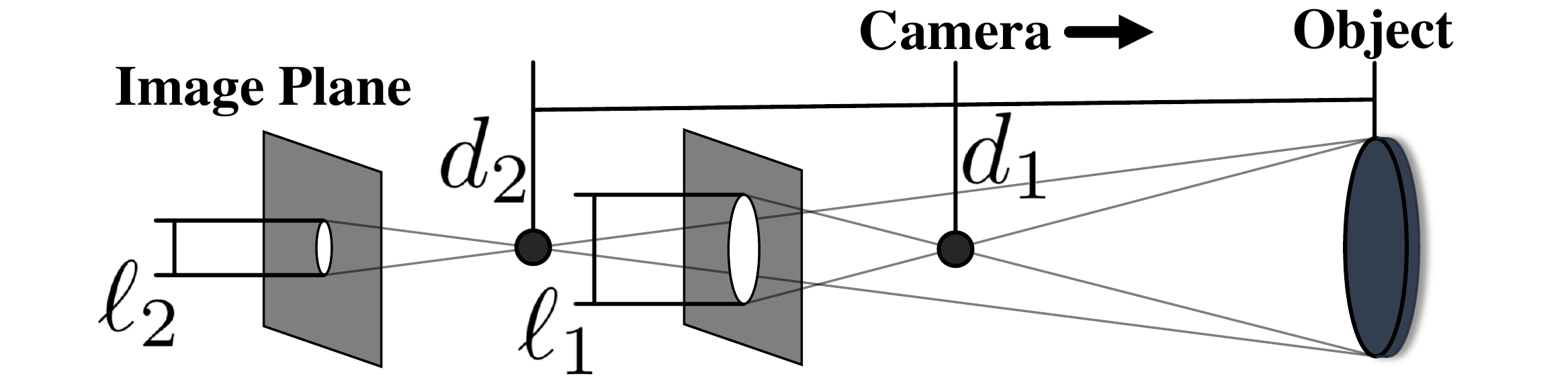}
	\caption{ \textbf{Optical Expansion and Depth.}
		An object's projection onto the image plane scales inversely with the depth between the camera and object.
		We determine an object's depth ($d$) using video object segmentation, relative camera movement, and corresponding changes in scale ($\ell$).
		In this example, $d_1 = \frac{d_2}{2}$, $\ell_1 = 2 \ell_2$, and $d_1 \ell_1 = d_2 \ell_2$
		%In this example, $d_1 = 2 d_2$, $\ell_1 = \frac{ \ell_2}{2}$, and $d_1 \ell_1 = d_2 \ell_2$
		%		In this example, $d_1 = 2 d_2 = 4 d_3$, $\ell_1 = \frac{ \ell_2}{2} = \frac{\ell_3}{4}$, and $d_1 \ell_1 = d_2 \ell_2 = d_3 \ell_3 $.%$\ell_1 d_1 = \ell_2 d_2 = \ell_3 d_3$.
	}
	\label{fig:optical_expansion}
\end{figure}

Given that many VOS methods perform well in unstructured environments, in this work, we show that VOS can similarly support 3D perception for robots and autonomous vehicles.
We take inspiration from work in psychology that establishes how people perceive depth motion from the optical expansion or contraction of objects  \cite{It51,SwGo86}, and we develop a deep network that learns object depth estimation from uncalibrated camera motion and video object segmentation (see Fig.~\ref{fig:front_vos}).
We depict our optical expansion model in Fig.~\ref{fig:optical_expansion}, which uses a moving pinhole camera and binary segmentation masks for an object in view.
To estimate an object's depth, we only need segmentations at two distances with an estimate of relative camera movement.
Notably, most autonomous hardware platforms already measure movement, and even hand-held devices can track movement using an inertial measurement unit or GPS.
Furthermore, although we do not study it here, if hardware-based measurements are not available, structure from motion is also plausible to recover camera motion \cite{KaGaBa19,MuTa17,ScFr16}.

In recent work \cite{GrFlCo20}, we use a similar model for VOS-based visual servo control, depth estimation, and mobile robot grasping.
However, our previous analytic depth estimation method does not adequately account for segmentation errors.
For real-world objects in complicated scenes, segmentation quality can change among frames, with typical errors including: incomplete object segmentation, partial background inclusion, or segmenting the wrong object.
Thus, we develop and train a deep network that learns to accommodate segmentation errors and reduces object depth estimation error from \cite{GrFlCo20} by as much as 59\%.

The first contribution of our paper is developing a learning-based approach to object depth estimation using motion and segmentation,  
which we experimentally evaluate in multiple domains.
To the best of our knowledge, this work is the first to use a learned, segmentation-based approach to depth estimation, which has many advantages.
First, we use segmentation masks as input, so our network does not rely on application-specific visual characteristics and is useful in multiple domains.
Second, we process a series of observations simultaneously, thereby mitigating errors associated with any individual camera movement or segmentation mask.
Third, our VOS implementation operates on streaming video and our method, using a single forward pass, runs in real-time.
Fourth, our approach only requires a single RGB camera and relative motion (no 3D sensors).
Finally, our depth estimation accuracy will improve with future innovations in VOS.

%{\color{note}[need to point out that we validate our approach in multiple domains]}

A second contribution of our paper is the \textbf{O}bject \textbf{D}epth via \textbf{M}otion and \textbf{S}egmentation (ODMS) dataset.\footnote{Dataset and source code website: \url{https://github.com/griffbr/ODMS}}
This is the first dataset for VOS-based depth estimation and enables learning-based algorithms to be leveraged in this problem space. 
ODMS data consist of a series of object segmentation masks, camera movement distances, and ground truth object depth. 
Due to the high cost of data collection and user annotation  \cite{DAVIS2018,ViGr09}, manually collecting training data would either be cost prohibitive or severely limit network complexity to avoid overfitting.
Instead, we configure our dataset to continuously generate synthetic training data with random distances, object profiles, and even perturbations, so we can train networks of arbitrary complexity.
%On the other hand, we can train networks of arbitrary complexity using our dataset to continuously-generate new training data with random distances, object profiles, and even perturbations.
Furthermore, because our network input consists simply of binary segmentation masks and distances, we show that domain transfer from synthetic training data to real-world applications is viable.
Finally, as a benchmark evaluation, we create four ODMS validation and test sets with 15,650 examples in multiple domains, including robotics and driving.
%We create four ODMS validation and test sets with over 15,650 examples in multiple domains, including robotics and driving.

\section{Related Work}

We use video object segmentation (VOS) to process raw input video and output the binary segmentation masks we use to estimate object depth in this work.
Unsupervised VOS usually relies on generic object motion and appearance cues \cite{NLC,GrCoWACV2019,KEY,FST,WeSz17,LIBSVX}, while semi-supervised VOS segments objects that are specified in user-annotated examples \cite{CINM,PML,GrCo19,PREMVOS,RGMP,OSMN}.
Thus, semi-supervised VOS can learn a specific object's visual characteristics and reliably segment dynamic \textit{or} static objects.
%To improve semi-supervised VOS, recent work focuses on automatic annotation frame-selection \cite{GrCo19} or using robots to generate training data \cite{FlCoGr19,SiEtAl19}.
To segment objects in our robot experiments, we use One-Shot Video Object Segmentation (OSVOS) \cite{OSVOS}.
OSVOS is state-of-the-art in VOS, has influenced other leading methods \cite{OSVOS-S,OnAVOS}, and does not require temporal consistency (OSVOS segments frames independently). 
During robot experiments, we apply OSVOS models that have been pre-trained with annotated examples of each object rather than annotating an example frame at inference time.
%For robot experiments, we cannot provide annotated examples at inference, so we pre-train OSVOS for each object using a set of previously-annotated frames.

We take inspiration from many existing datasets in this work.
VOS research has benefited from benchmark datasets like SegTrackv2 \cite{SegTrack,SegTrackv2}, DAVIS \cite{DAVIS,DAVIS17}, and YouTube-VOS \cite{YTVOS}, which have provided increasing amounts of annotated training data.
The recently developed MannequinChallenge dataset \cite{LiEtAl19} trained a network to predict dense depth maps from videos with people, with improved performance when given an additional human-mask input.
Among automotive datasets, Cityscapes \cite{City} focuses on \textit{semantic segmentation} (i.e., assigning class labels to all pixels), KITTI \cite{KITTI} includes benchmarks separate from segmentation for single-image depth completion and prediction, and SYNTHIA \cite{SYNTHIA} has driving sequences with simultaneous ground truth for semantic segmentation and depth images.
In this work, our ODMS dataset focuses on \textbf{O}bject \textbf{D}epth via \textbf{M}otion and \textbf{S}egmentation, establishing a new benchmark for segmentation-based 3D perception in robotics and driving.
In addition, ODMS is arbitrarily extensible, which makes learning-based methods feasible in this problem space.

\section{Optical Expansion Model}
\label{sec:solve}

Our optical expansion model (Fig.~\ref{fig:optical_expansion}) forms the theoretical underpinning for our learning-based approach in Section~\ref{sec:learn} and ODMS dataset in Section~\ref{sec:dataset}.
In this section, we derive the complete model and analytic solution for segmentation-based depth estimation. %, including the solution used in \cite{GrFlCo20}.
We start by defining the inputs we use to estimate depth.
Assume we are given a set of $n \geq 2$ observations that consist of masks 
\begin{align}
	\mathsf{M} := \{\mathbf{M}_1, \mathbf{M}_2, \cdots, \mathbf{M}_n\}
	%	\mathsf M := \{\boldsymbol{M}_1, m_2, \cdots, m_n\}
	\label{eq:masks}
\end{align}
segmenting an object and corresponding camera positions on the optical axis
\begin{align}
	\mathbf{z} := \{z_1, z_2, \cdots, z_n\}.
	%	Z := \{z_1, z_2, \cdots, z_n\}.
	\label{eq:distances}
\end{align}
Each binary mask image $\mathbf{M}_i$ consists of pixel-level labels where 1 indicates a pixel belongs to a specific segmented object and 0 is background. %$\boldsymbol{M}$
For the solutions in this work, the optical axis's origin and absolute position of  $\mathbf{z}$ is inconsequential.

\subsection{Relating Depth and Scale}

We use changes in scale of an object's segmentation mask to estimate depth.
As depicted in Fig.~\ref{fig:optical_expansion}, we relate depth and scale across observations using
\begin{align}
	d_i \ell_i = d_j \ell_j \implies \frac{\ell_j}{\ell_i} = \frac{d_i}{d_j}, % \implies \sqrt{A_1} d_1 = \sqrt{A_2} d_2,
	\label{eq:elld}
\end{align}
where $\ell_i$ is the object's projected scale in $\mathbf{M}_i$, $d_i$ is the distance on the optical axis from $z_i$ to the visible perimeter of the segmented object, and $\frac{\ell_j}{\ell_i}$ is the object's change in scale between $\mathbf{M}_i$ and $\mathbf{M}_j$.
Notably, it is more straightforward to track changes in scale using area (i.e., the sum of mask pixels) than length measurements.
Thus, we use Galileo Galilei's Square-cube law to modify \eqref{eq:elld} as
%It is more practical to measure area across observations than $\ell$, so we use Galileo Galilei's Square-cube law to modify \eqref{eq:elld} as
\begin{align}
	a_j = a_i \bigg(\frac{\ell_j}{\ell_i}\bigg)^2 \implies
	\frac{\ell_j}{\ell_i} = \frac{\sqrt{a_j}}{\sqrt{a_i}} = \frac{d_i}{d_j}, % \implies \sqrt{A_1} d_1 = \sqrt{A_2} d_2,
	\label{eq:galileo}
\end{align}
where $a_i$ is an object's projected area at $d_i$ and $\frac{\sqrt{a_j}}{\sqrt{a_i}}$ is equal to the change in scale between $\mathbf{M}_i$ and $\mathbf{M}_j$.
%Notably, we measure $A$ by using the sum of object segmentation pixels.
Combining \eqref{eq:elld} and \eqref{eq:galileo}, we relate observations as %to constant $c$ as
\begin{align}
	d_i \sqrt{a_i} = d_j \sqrt{a_j} = c,
	\label{eq:c}
\end{align}
where $c$ is a constant corresponding to an object's orthogonal surface area.

\begin{figure}[t]
	\centering
	\includegraphics[width=0.75\textwidth]{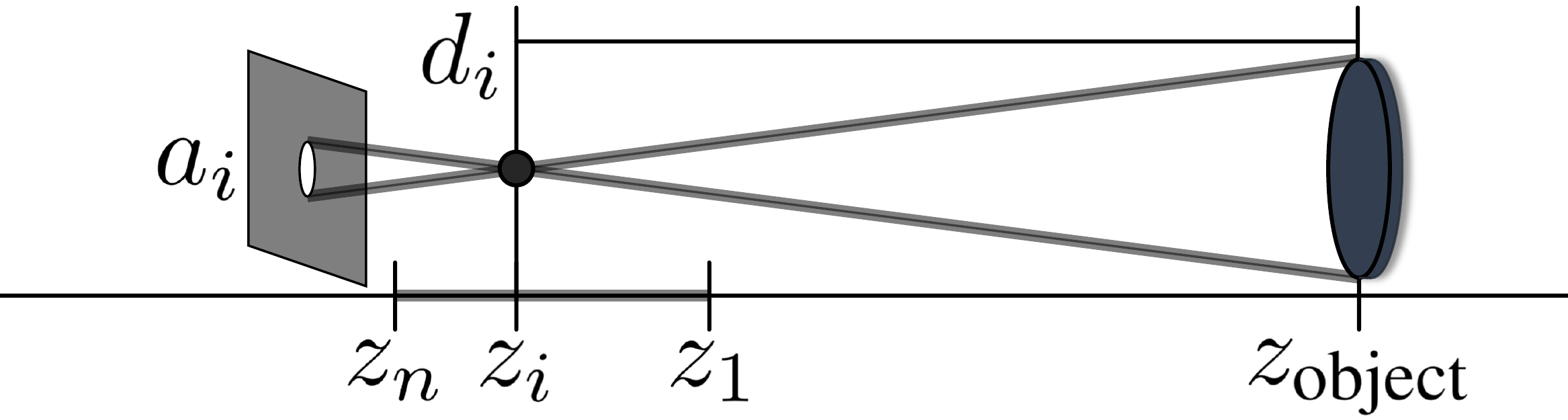}
	%	\caption{ \textbf{Solving for Object Depth.}
	\caption{ \textbf{Calculating Object Depth.}
		%		To calculate depth $d_i$, we relate the $z$-axis coordinate $z_{\text{object}}$ to moving camera pose $z_i$.
		First, we define $d_i$ in terms of its component parts $z_{\text{object}}$ and $z_i$ \eqref{eq:di}.
		Second, we relate measured changes in camera pose $z_i$ and segmentation area $a_i$ across observations \eqref{eq:cobj}.
		Finally, we solve for $z_{\text{object}}$ using \eqref{eq:zobj}% or \eqref{eq:axb}}
	}
	\label{fig:optical_expansion_za}
\end{figure}

\subsection{Object Depth Solution}

To find object depth $d_i$ in \eqref{eq:c}, we first redefine $d_i$ in terms of its components as
%To solve \eqref{eq:c} for object depth $d_i$, we define $d_i$ as
\begin{align}
	d_i := z_i - z_{\text{object}},
	\label{eq:di}
\end{align}
where $z_{\text{object}}$  is the object's static position on the optical axis and $\dot{z}_{\text{object}}=0$ (see Fig.~\ref{fig:optical_expansion_za}).
Substituting \eqref{eq:di} in \eqref{eq:c}, we can now relate observations as
\begin{align}
	(z_i - z_{\text{object}})  \sqrt{a_i} = (z_j  - z_{\text{object}}) \sqrt{a_j} = c.
	% (z_{\text{object}} - z_i)  \sqrt{A_i} =   (z_{\text{object}} - z_j) \sqrt{A_j} = c.
	\label{eq:cobj}
\end{align}
From \eqref{eq:cobj}, we can solve $z_{\text{object}}$ from any two unique observations ($z_i \neq z_j$) as
%From any two observations, we can solve \eqref{eq:cobj} for $z_{\text{object}}$ as
\begin{align}
	z_{\text{object}} = \frac{z_i \sqrt{a_i} - z_j \sqrt{a_j}}{\sqrt{a_i} - \sqrt{a_j}} = \frac{z_i  - z_j \frac{\sqrt{a_j}}{\sqrt{a_i}}}{1 - \frac{\sqrt{a_j}}{\sqrt{a_i}}}.
	\label{eq:zobj}
\end{align}
Substituting $z_{\text{object}}$ in \eqref{eq:di}, we can now find object depth $d_i$ at any observation.

\section{Learning Object Depth from Camera Motion and Video Object Segmentation}
%\section{Learning to Estimate Depth using Camera Motion and Video Object Segmentation}
\label{sec:learn}

Using the optical expansion model from Section~\ref{sec:solve}, we design a deep network, \textbf{O}bject \textbf{D}epth \textbf{N}etwork (ODN),  
that learns to predict the depth of segmented objects given a series of binary masks $\mathsf{M}$ \eqref{eq:masks} and changes in camera position $\mathbf{z}$ \eqref{eq:distances}.
%We call this network \textbf{D}epth estimation via \textbf{O}ptical \textbf{E}xpansion of \textbf{S}egmentation \textbf{Net}work (DOESNet).
To keep ODN broadly applicable, we formulate a normalized relative distance input in Section~\ref{sec:norm}.
In Sections~\ref{sec:dist} and \ref{sec:scale}, we derive three unique losses for learning depth estimation. % using changes in relative scale between segmentation masks.
After some remarks on using intermediate observations in Section~\ref{sec:intermediate}, we detail our ODN architecture in Section~\ref{sec:net}.

\subsection{Normalized Relative Distance Input}
\label{sec:norm}
%\subsection{Normalized Distance and Other Inputs}

To learn to estimate a segmented object's depth, we first derive a normalized relative distance input that increases generalization.
As in Section~\ref{sec:solve}, assume we are given a set of $n$ segmentation masks $\mathsf{M}$ with corresponding camera positions $\mathbf{z}$.
We can use $\mathsf{M}$ and $\mathbf{z}$ as inputs to predict object depth, however, a direct $\mathbf{z}$ input enables a learned prior based on absolute camera position, which limits applicability at inference. 
To avoid this, we define a relative distance input
\begin{align}
	\Delta  \mathbf{z} := \{z_2 - z_1, z_3 - z_1, \cdots, z_n - z_1\},
	%	\Delta Z := \{z_2 - z_1, z_3 - z_1, \cdots, z_n - z_1\},
	\label{eq:rel}
\end{align}
where $z_1, z_2, \cdots, z_n$ are the sorted $\mathbf{z}$ positions with the minimum $z_1$ closest to the object (see Fig.~\ref{fig:optical_expansion_za}) and $\Delta  \mathbf{z} \in \mathbb{R}^{n-1}$. 
Although $\Delta  \mathbf{z}$ consists only of relative changes in position, it still requires learning a specific SI unit of distance and enables a prior based on camera movement range.
Thus, we normalize \eqref{eq:rel} as %$\Delta Z$ as
\begin{align}
	\mathbf{\bar{z}} := \Big\{\frac{z_i - z_1}{z_n - z_1} | z \in \mathbf{z}, 1 < i < n \Big\},
	\label{eq:znorm}	
\end{align}
where $z_n - z_1$ is the camera move range, $\frac{z_i - z_1}{z_n - z_1} \in (0,1)$, and $\mathbf{\bar{z}} \in \mathbb{R}^{n-2}$. %, and $\Delta \bar{Z} \in \mathbb{R}^{n-2}$.

Using $\mathbf{\bar{z}}$ as our camera motion input increases the general applicability of ODN.
First, $\mathbf{\bar{z}}$ uses the relative difference formulation, so ODN does not learn to associate depth with an absolute camera position.
Second, $\mathbf{\bar{z}}$ is dimensionless, so our trained ODN can use camera movements on the scale of millimeters or kilometers (it makes no difference). 
Finally, $\mathbf{\bar{z}}$ is made a more compact motion input by removing the unnecessary constants $\frac{z_1 - z_1}{z_n - z_1} = 0$ and $\frac{z_n - z_1}{z_n - z_1} = 1$ in \eqref{eq:znorm}.

%Also limits complexity of the input since $z_{(1,1)} = 0$ and $z_{(1,n)} = 1$ are constants that can be learned without an explicit input.
%$A_i d_i z_i A_j d_j z_j z_{\text{object}}$

\subsection{Normalized Relative Depth Loss}
\label{sec:dist}

Our basic depth loss, given input masks $\mathsf{M}$ \eqref{eq:masks} and relative distances $\Delta  \mathbf{z}$ \eqref{eq:rel}, is
\begin{align}
	\mathcal{L}_d(\textbf{W}) := d_1 - f_d(\mathsf{M}, \Delta  \mathbf{z}, \mathbf{W}),
%	\mathcal{L}_d(\textbf{W}) := \left| d_1 - f_d(\mathsf{M}, \Delta  \mathbf{z}, \mathbf{W})  \right|,
	\label{eq:depthloss}
\end{align}
where $\textbf{W}$ are the trainable network parameters, $d_1$ is the ground truth object depth at $z_1$ \eqref{eq:di}, and $f_d \in \mathbb{R}$ is the predicted depth.
To use the normalized distance input $\mathbf{\bar{z}}$ \eqref{eq:znorm}, we modify \eqref{eq:depthloss} and define a normalized depth loss as
%As motivated in Section~\ref{sec:norm}, using normalized distance input $\Delta \bar{Z}$ \eqref{eq:znorm} enables broader application. Thus, we modify \eqref{eq:depthloss} to define a normalized-distance loss
\begin{align}
	%	\mathcal{L}_z(\textbf{W}) := \left| \bar{z}_{\text{object}} - f(M, \Delta \bar{Z}, \mathbf{W}) \right|,
	%\\
	%	\mathcal{L}_z(\textbf{W}) := \left| z_{\text{object}} - f(M, \Delta \bar{Z}, \mathbf{W}) (z_n - z_1) \right|,
	%	\\
	%		\mathcal{L}_z(\textbf{W}) := \left| z_{\text{object}} - (z_n - z_1) f(M, \Delta \bar{Z}, \mathbf{W}) \right|,
	%	\\
	\mathcal{L}_{\bar{d}}(\textbf{W}) := \frac{d_1}{z_n - z_1} - f_{\bar{d}}(\mathsf{M}, \mathbf{\bar{z}}, \mathbf{W}),
	%	\mathcal{L}_{\bar{z}}(\textbf{W}) := \left| z_{\text{object}} - f(M, \Delta \bar{Z}, \mathbf{W}) (z_n - z_1) \right|,
	\label{eq:normloss}
\end{align} 
where $\frac{d_1}{z_n - z_1}$ is the normalized object depth and $f_{\bar{d}}$ is a dimensionless depth prediction that is in terms of the input camera movement range.
To use $f_{\bar{d}}$ at inference, we multiply the normalized output $f_{\bar{d}}$ by $(z_n - z_1)$ to find $d_1$.

\subsection{Relative Scale Loss}
\label{sec:scale}

We increase depth accuracy and simplify ODN's prediction by learning to estimate relative changes in segmentation scale.
In Section~\ref{sec:dist}, we define loss functions that use a similar input-output paradigm to the analytic solution in Section~\ref{sec:solve}.
However, training ODN to directly predict depth requires learning many operations.
Alternatively, if ODN only predicts the relative change in segmentation scale, we can finish calculating depth using \eqref{eq:zobj}.
Thus, we define a loss for predicting the relative scale as
\begin{align}
	\mathcal{L}_{\ell}(\textbf{W}) := \frac{\ell_n}{\ell_1} - f_\ell(\mathsf{M}, \mathbf{\bar{z}}, \mathbf{W}),
	%	\mathcal{L}_{\ell}(\textbf{W}) := \left| \frac{\ell_n}{\ell_1} - f(\mathsf{M}, \mathbf{\bar{z}}, \mathbf{W})  \right|,
	\label{eq:scaleloss}
\end{align}
where $\frac{\ell_n}{\ell_1}=\frac{d_1}{d_n} \in (0,1)$ \eqref{eq:elld} is the ground truth distance-based change in scale between $\mathbf{M}_n$ and $\mathbf{M}_1$ and $f_\ell$ is the predicted scale change.
To use $f_\ell$ at inference, we output $f_\ell \approx \frac{\ell_n}{\ell_1}$ and, using \eqref{eq:galileo} to substitute $\frac{\ell_j}{\ell_i}$ for $\frac{\sqrt{a_j}}{\sqrt{a_i}}$ in \eqref{eq:zobj}, find $z_{\text{object}}$ as
\begin{align}
	z_{\text{object}} =  \frac{z_1  - z_n f_\ell }{1 - f_\ell } \approx \frac{z_1  - z_n \big( \frac{\ell_n}{\ell_1} \big) }{1 - \big( \frac{\ell_n}{\ell_1} \big) }.
	%z_{\text{object}} = \frac{z_1  - z_n \big( \frac{\ell_n}{\ell_1} \big) }{1 - \big( \frac{\ell_n}{\ell_1} \big) } \approx \frac{z_1  - z_n f }{1 - f },
	\label{eq:zobjscale}
\end{align}
%where $f \approx \frac{\ell_n}{\ell_1}=\frac{\sqrt{A_n}}{\sqrt{A_1}}$ replaces relative scale $ \frac{\sqrt{A_j}}{\sqrt{A_i}}$ in \eqref{eq:zobj}.
After finding $z_{\text{object}}$ in \eqref{eq:zobjscale}, we use \eqref{eq:di} to find object depth as $d_1 = z_1 - z_\text{object}$.

\subsection{Remarks on using Intermediate Observations}
\label{sec:intermediate}

Although the ground truth label $d_1$ in \eqref{eq:depthloss}-\eqref{eq:normloss} is determined only by camera position $z_1$ and label $\frac{\ell_n}{\ell_1}$ in \eqref{eq:scaleloss} is determined only by endpoint masks $\mathbf{M}_n$, $\mathbf{M}_1$, we emphasize that intermediate mask and distance inputs are still useful.
Consider that, first, the ground truth mask scale monotonically decreases across all observations (i.e., $\forall i, \ell_{i+1} < \ell_i$).
Second, the distance inputs make it possible to extrapolate $d_1$ and $\frac{\ell_n}{\ell_1}$ from intermediate changes in scale.
Third, if $z_1$, $z_n$, $\mathbf{M}_1$, or $\mathbf{M}_n$ have significant errors, intermediate observations provide the best prediction for $d_1$  or  $\frac{\ell_n}{\ell_1}$.
Finally, experiments in Section~\ref{sec:exp_n} show that intermediate observations improve performance for networks trained on \eqref{eq:depthloss}, \eqref{eq:normloss}, or \eqref{eq:scaleloss}.

\subsection{Object Depth Estimation Network Architecture}
\label{sec:net}

\begin{figure}[t]
	\centering
	\includegraphics[width=0.975\textwidth]{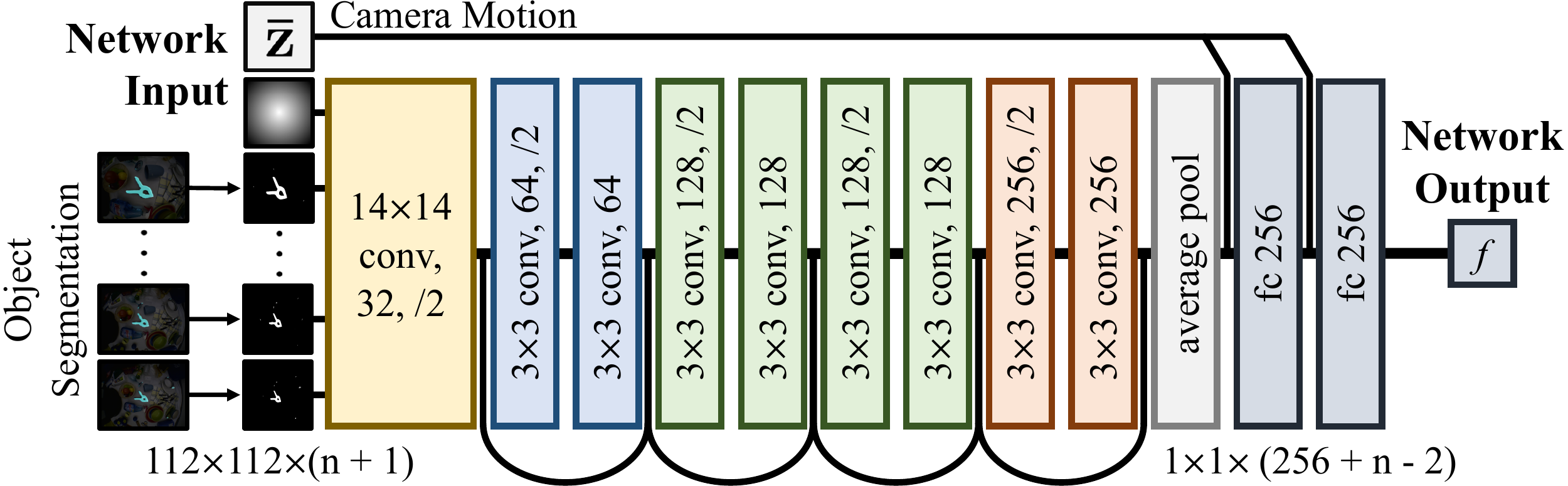}
	\caption{ \textbf{Object Depth Network Architecture} 
	}
	\label{fig:network}
\end{figure}

Our ODN architecture is shown in Fig.~\ref{fig:network}.
The input to the first convolution layer consists of $n$ 112$\times$112 binary segmentation masks and, for three configurations in Section~\ref{sec:radial}, a radial image.
The first convolution layer uses 14$\times$14 kernels, and the remaining convolution layers use 3$\times$3 kernels in four residual blocks \cite{HeEtAl16}.
After average pooling the last residual block, the relative camera position (e.g., $\mathbf{ \bar{z}}$) is included with the input to the first two fully-connected layers, which use ReLU activation and 20\% dropout for all inputs during training \cite{dropout14}.
After the first two fully-connected layers, our ODN architecture ends with one last fully-connected neuron that, depending on chosen loss, is the output object depth $f_d(\mathsf{M}, \Delta \mathbf{z}, \mathbf{W}) \in \mathbb{R}$ using \eqref{eq:depthloss}, normalized object depth $f_{\bar{d}}(\mathsf{M}, \mathbf{\bar{z}}, \mathbf{W})$ using \eqref{eq:normloss}, or relative scale $f_\ell(\mathsf{M}, \mathbf{\bar{z}}, \mathbf{W})$ using \eqref{eq:scaleloss}.

\section{ODMS Dataset}
%\section{Object Depth via Motion and Segmentation ODMS Dataset}
\label{sec:dataset}

To train our object depth networks from Section~\ref{sec:learn}, we introduce the \textbf{O}bject \textbf{D}epth via \textbf{M}otion and \textbf{S}egmentation dataset (ODMS).
In Section~\ref{sec:gen_masks}, we explain how ODMS continuously generates new labeled training data, making learning-based techniques feasible in this problem space.
In Section~\ref{sec:sets}, we describe the robotics-, driving-, and simulation-based test and validation sets we develop for evaluation.
Finally, in Section~\ref{sec:train}, we detail our ODMS training implementation.

\subsection{Generating Random Object Masks at Scale}
\label{sec:gen_masks}

\subsubsection{Camera Distance and Depth}
We generate new training data by, first, determining $n$ random camera distances (i.e., $\mathbf{z}$ \eqref{eq:distances}) for each training example.
To make ODMS configurable, assume we are given a minimum camera movement range ($\Delta z_{\text{min}}$) and minimum and maximum object depths ($d_{\text{min}}$, $d_{\text{max}}$). 
Using these parameters, we define distributions for uniform random variables to find the endpoints
\begin{align}
	\label{eq:z1}
	z_1 \sim & ~\mathcal{U}[d_{\text{min}},d_{\text{max}}-\Delta z_{\text{min}}], \\
	z_n \sim & ~\mathcal{U}[z_1 + \Delta z_{\text{min}},d_{\text{max}}],
	\label{eq:zn}
\end{align}
and, for $1 < i < n$, the remaining intermediate camera positions 
\begin{align}
	z_i \sim \mathcal{U}(z_1, z_n).
	\label{eq:zi}
\end{align}
Using \eqref{eq:z1}-\eqref{eq:zi} to select $\mathbf{z} = \{z_1, \cdots, z_n\}$ ensures that the random camera movement range is independent of the number of observations $n$.
For the object depth label $d_1$, we choose an optical axis such that $z_\text{object}=0$ and $d_1=z_1$ \eqref{eq:di}.
We generate data in this work using $\Delta z_\text{min}=d_\text{min}=0.1~\textrm{m}$ and $d_\text{max}=0.7~\textrm{m}$.

\begin{figure}[t]
	\centering
	\includegraphics[width=0.975\textwidth]{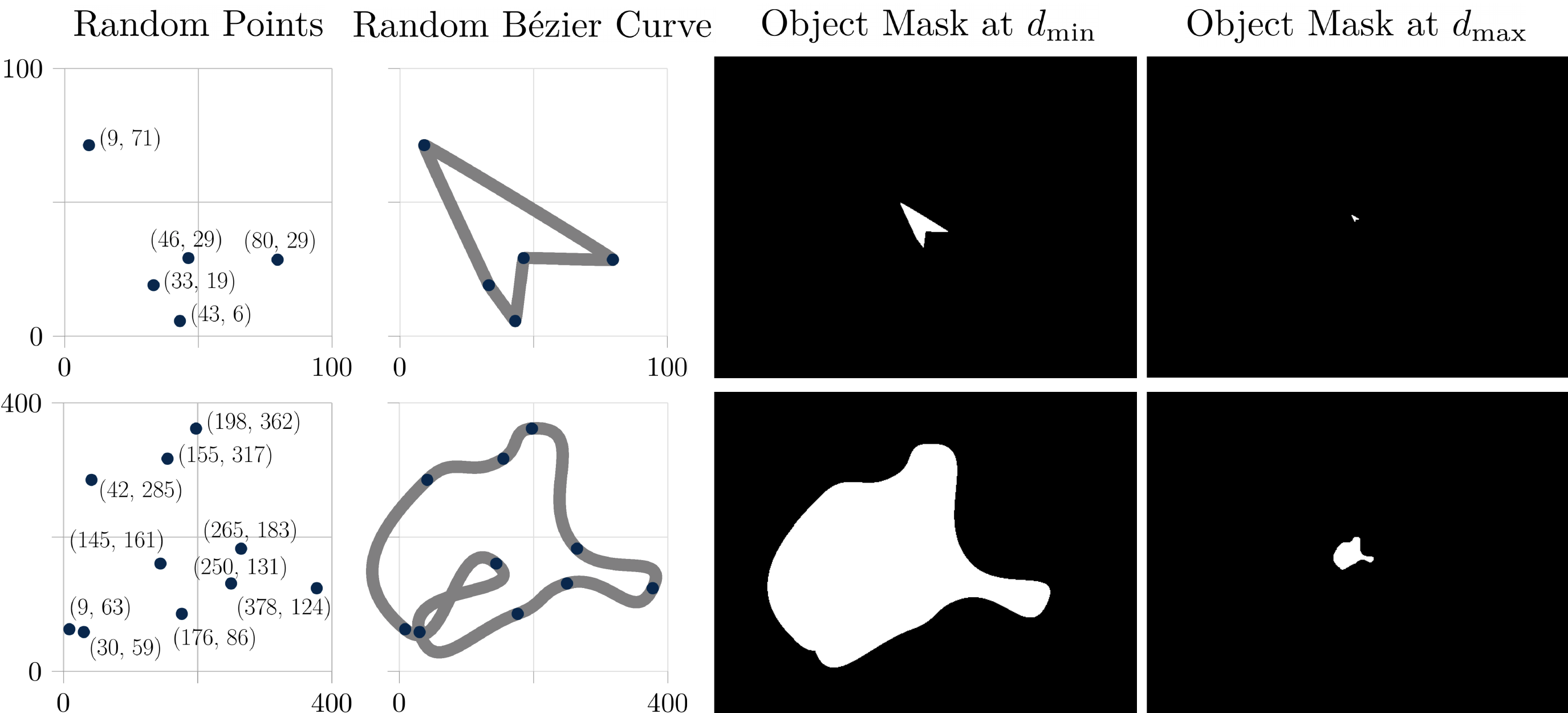}
	\caption{ \textbf{Generating Random Object Masks at Scale.} 
		Initializing from a random number of points within a variable boundary (\textit{left}), random curves complete the contour of each simulated object (\textit{middle left}).
		%	Generated objects start as random points within varying bounds (e.g., 100 (\textit{top left}) or 400 (\textit{bottom left})).
		%	Next, B\'ezier curves with randomly selected smoothness complete the contour of each simulated object (\textit{middle left}).
		These contours are then scaled for each simulated distance and output as a filled binary mask (\textit{right}).
		Each generated object is unique
		%{\color{note} Mention how B\'ezier Curve can extend outside of point scale boundary.}
	}
	\label{fig:data_gen}
\end{figure}

\subsubsection{Random Object Contour and Binary Masks}
After determining $\mathbf{z}$, we generate a random object with $n$ binary masks (i.e., $\mathsf{M}$ \eqref{eq:masks}) scaled for each distance in $\mathbf{z}$ (see Fig.~\ref{fig:data_gen}).
To make each object unique, we randomly select parameters that change the object's size ($s_\mathbf{p}$), number of contour points ($n_\mathbf{p}$), and contour smoothness ($r_B$, $\rho_B$).
In this work, we randomly select $s_\mathbf{p}$ from $\{100, 200, 300, 400\}$ and $n_\mathbf{p}$ from $\{3, 4, \cdots, 10\}$.
Using $s_\mathbf{p}$ and $n_\mathbf{p}$, we select each of the random initial contour points, $\mathbf{p}_i \in \mathbb{R}^2$ for $1 \leq i \leq n_\mathbf{p}$, as
\begin{align}
	\mathbf{p}_i= [ x_i,y_i ]',~
	x_i \sim \mathcal{U}[0,s_\mathbf{p}],~ y_i \sim \mathcal{U}[0,s_\mathbf{p}].
	\label{eq:xy}
\end{align}

To complete the object's contour, we use cubic B\'ezier curves with random smoothness to connect each set of adjacent coordinates $\mathbf{p}_i$, $\mathbf{p}_j$ from \eqref{eq:xy}.
Essentially, $r_B$ and $\rho_B$ determine polar coordinates for the two intermediate B\'ezier control points of each curve. 
$\arctan (\rho_B)$ is the rotation of a control point away from the line connecting $\mathbf{p}_i$ and $\mathbf{p}_j$,
while $r_B $ is the relative radius of a control point away from $\mathbf{p}_i$ (e.g., $r_B=1$ has a radius of $\lVert \mathbf{p}_i - \mathbf{p}_j \rVert$).
In this work, we randomly select $r_B$ from $\{0.01, 0.05, 0.2, 0.5\}$ and $\rho_B$ from $\{0.01, 0.05, 0.2\}$ for each object.
In general, lower $r_B$ and $\rho_B$ values result in a more straight-edged contour, while higher values result in a more curved and widespread contour.
As two illustrative examples in Fig.~\ref{fig:data_gen}, the top ``straight-edged" object uses $r_B=\rho_B=0.01$ and the bottom ``curved" object uses $r_B=0.5$ and $\rho_B=0.2$.

To simulate object segmentation over multiple distances, we scale the generated contour to match each distance $z_i \in \mathbf{z}$ from \eqref{eq:z1}-\eqref{eq:zi} and output a set of binary masks $\mathbf{M}_i \in \mathsf{M}$ \eqref{eq:masks}.
We let the initial contour represent the object's image projection at $d_{\text{min}}$, and designate this initial scale as $\ell_{\text{min}}=1$.
Having chosen an optical axis such that $z_\text{object}=0$ in \eqref{eq:di} (i.e., $d_i=z_i$), we modify \eqref{eq:elld} to find the contour scale of each mask,  $\ell_i $ for $1 \leq i \leq n$, as
\begin{align}
	\ell_i = \frac{d_\text{min} \ell_\text{min}}{d_i}= \frac{d_\text{min}}{z_i}.
	\label{eq:gen_scale}
\end{align}
After finding $\ell_i$, we scale, fill, and add the object contour to each mask $\mathbf{M}_i$.
In this work, we position the contour by centering the scaled boundary ($\ell_i s_\mathbf{p}$) in a 480$\times$640 mask.
Our complete object-generating process is shown in Fig.~\ref{fig:data_gen}.

\subsection{Robotics, Driving, and Simulation Validation and Test Sets}
\label{sec:sets}

\begin{figure}[t]
	\centering
	\includegraphics[width=0.975\textwidth]{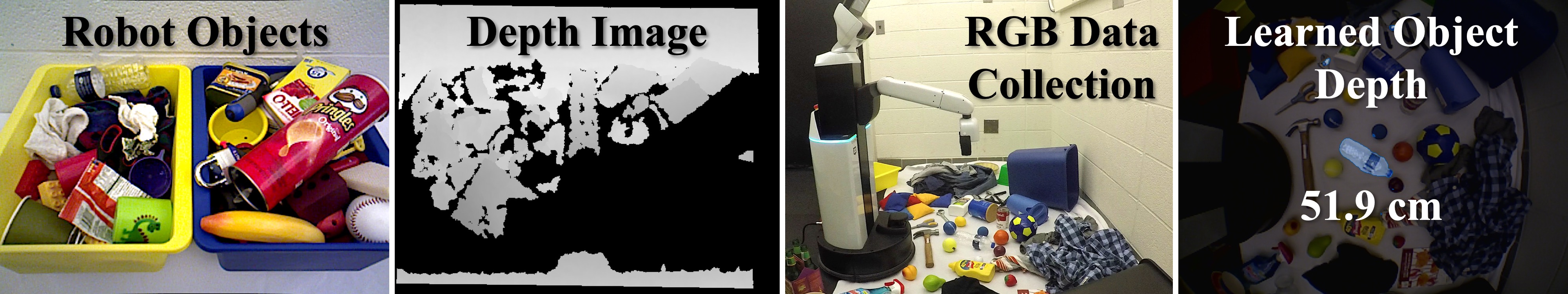}
	\caption{ \textbf{Robot Experiment Data.}
		HSR view of validation (yellow bin) and test set objects (blue bin) using head-mounted RGBD camera (\textit{left}).
		Unfortunately, the depth image is missing many objects (\textit{middle left}).
		However, using 4,400 robot-collected examples (\textit{middle right}), we find that segmentation-based object depth works (\textit{right})
	}
	\label{fig:val_test}
\end{figure}

%we make an effort to generate repeatable experimentation practices, but it is not always plausible for robotics. So, we do multiple experiment setups...

We test object depth estimation in a variety of settings using four ODMS validation and test sets.
These are based on robot experiments, driving, and simulated data with and without perturbations and provide a repeatable benchmark for ablation studies and future methods.
All examples include $n \geq 10$ observations.

%All sets use $n \geq 10$ observations per labeled example.

\subsubsection{Robot Validation and Test Set}
Our robot experiment data provide an evaluation for object depth estimation from a physical platform using video object segmentation on real-world objects in a practical use case.
We collect data using a Toyota Human Support Robot (HSR), which has a 4-DOF manipulator arm with an end effector-mounted wide-angle grasp camera \cite{UiYamaguchi2015,HSR_journal}.
Using HSR's prismatic torso, we collect 480$\times$640 grasp-camera images as the end effector approaches an object of interest, with the intent that HSR can estimate the object's depth using motion and segmentation.
We use 16 custom household objects for our validation set and 24 YCB objects \cite{YCB} for our test set (Fig.~\ref{fig:val_test}, left).
For each object, we collect 30 images distanced 2~\textrm{cm} apart of the object in isolation and, as an added challenge, 30 more images in a cluttered setting (see Fig.~\ref{fig:val_test}, middle right).
The ground truth object depth ($d_1$) is manually measured at the closest camera position and propagated to the remaining images using HSR's kinematics and encoder values, which also measure camera positions ($\mathbf{z}$).
To generate binary masks ($\mathsf{M}$), we segment objects using OSVOS \cite{OSVOS}, which we fine-tune on each object using three annotated images from outside of the validation and test sets.
We vary the input camera movement range between 18-58~\textrm{cm} and object depth ($d_1$) between 11-60~\textrm{cm} to generate 4,400 robot object depth estimation examples (1,760 validation and 2,640 test).

\subsubsection{Driving Validation and Test Set}
\begin{figure}[t]
	\centering
	\includegraphics[width=0.975\textwidth]{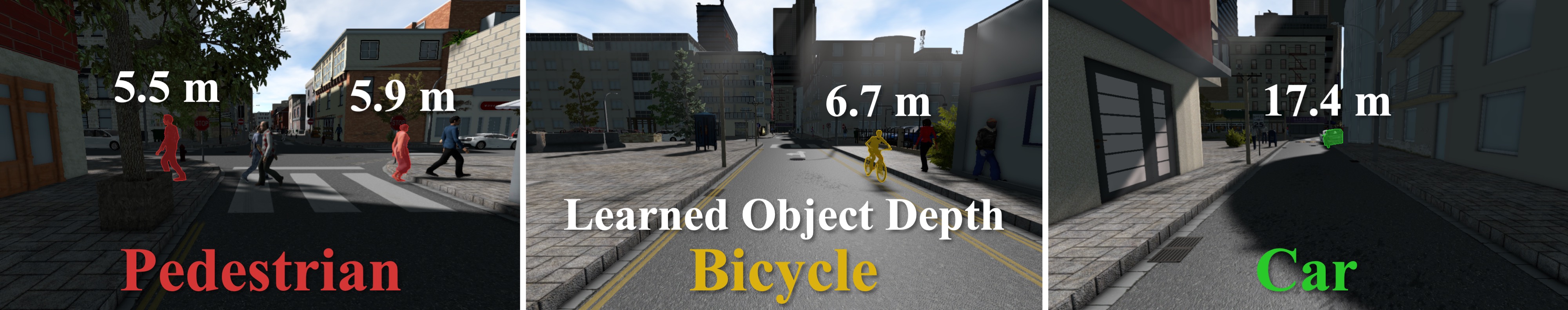}
	\caption{ \textbf{Driving Test Set Examples and Results.}
		The ODN$_\ell$ object depth error is -6 and -23~\textrm{cm} for the pedestrians, -10~\textrm{cm} for the bicycle, and -4~\textrm{cm} for the car
		%		The predicted object depth error is -6, -23~\textrm{cm} for the pedestrians, -10~\textrm{cm} for the bicycle, and -4~\textrm{cm} for the car
		%Depth error is -0.06, -0.23, -0.10, and -0.04~\textrm{m} from left to right.
	}
	\label{fig:val_synthia}
\end{figure}

Our driving data provide an evaluation for object depth estimation in a faster moving automotive domain with greater camera movement and depth distances.
Our goal is to track driving obstacles using an RGB camera, segmentation, and vehicle odometry.
Challenges include changing object perspectives, camera rotation from vehicle turning, and moving objects.
We collect data using the SYNTHIA Dataset \cite{SYNTHIA}, which includes ground truth semantic segmentation, depth images, and vehicle odometry in a variety of urban scenes and weather conditions.
To generate binary masks ($\mathsf{M}$), we use SYNTHIA's semantic segmentation over a series of 760$\times$1280 frames for unique instances of pedestrians, bicycles, and cars (see Fig.~\ref{fig:val_synthia}).
For each instance, the ground truth object depth ($d_1$) is the mean depth image values contained within corresponding mask  $\mathbf{M}_1$.
As the vehicle moves, we track changes in camera position ($\mathbf{z}$) along the optical axis of position $z_1$.
With an input camera movement range between 4.2-68~\textrm{m} and object depth ($d_1$) between 1.5-62~\textrm{m}, we generate 1,250 driving object depth estimation examples (500 validation and 750 test).

\subsubsection{Simulation Validation and Test Sets}
\label{sec:sim_test}
Finally, we generate a set of normal and perturbation-based data for simulated objects. % using our mask-generation framework.
The normal set and the continuously-generated training data we use in Section~\ref{sec:train} both use the same mask-generating procedure from Section~\ref{sec:gen_masks}, so the normal set provides a consistent evaluation for the type of simulated objects we use during training.

To test robustness for segmentation errors, we also generate a set of simulated objects with random perturbations added to each mask, $\mathbf{M}_i$ for $1 \leq i \leq n$, as
\begin{align}
	{p}_i \sim  \mathcal{N}(0,1), 
	~\mathbf{M}_{i,{p}} = \begin{cases}
		\text{dilate}(\mathbf{M}_i, \lfloor {p}_i + 0.5 \rfloor) & \text{if}  ~ {p}_i \geq 0\\
		\text{erode}(\mathbf{M}_i, \lfloor {p}_i + 0.5 \rfloor) & \text{if}  ~ {p}_i < 0\\
	\end{cases},
	\label{eq:perturb}
\end{align}
where $\mathcal{N}(0,1)$ is a Gaussian distribution with $\mu=0$, $\sigma^2=1$, ${p}_i$ randomly determines the perturbation type and magnitude, and $\mathbf{M}_{i,{p}}$ is the perturbed version of initial mask $\mathbf{M}_i$.
Notably, the sign of ${p}_i$ determines a dilation or erosion perturbation, and the rounded magnitude of ${p}_i$ determines the number of iterations using a square connectivity equal to one.
When generating perturbed masks $\mathbf{M}_{i,{\rm p}}$, we make no other changes to input data or ground truth labels.

We generate 5,000 object depth estimation examples (2,000 validation and 3,000 test) for both the normal and perturbation-based simulation sets.

\subsection{Training Object Depth Networks using ODMS}
\label{sec:train}

Using the architecture in Section~\ref{sec:net}, we train networks for depth loss ${\mathcal{L}_{d}}$ \eqref{eq:depthloss}, normalized relative depth loss $\mathcal{L}_{\bar{d}}$  \eqref{eq:normloss}, and relative scale loss $\mathcal{L}_{\ell}$ \eqref{eq:scaleloss}.
We call these networks ODN$_d$, ODN$_{\bar{d}}$, and ODN$_\ell$ respectively.
We train each network with a batch size of 512 randomly-generated training examples using the framework in Section~\ref{sec:gen_masks} with $n=10$ observations per prediction. We train each network for 5,000 iterations using the Adam Optimizer \cite{adam14} with a $1\times10^{-3}$ learning rate, which takes 2.6 days using a single GPU (GTX 1080 Ti).
Notably, the primary time constraint for training is generating new masks, and we can train a similar configuration with $n=2$ for 5,000 iterations in 15 hours.

\section{Experimental Results}
\label{sec:results}

Our primary experiments and analysis use the four ODMS test sets.
For each test set, the number of network training iterations is determined by the best validation performance, which we check at every ten training iterations.
We determine the effectiveness of each depth estimation method using the mean percent error for each test set, which is calculated for each example as
\begin{align}
	\text{Percent Error} = \left| \frac{d_1 - \hat{d}_1}{d_1} \right| \times 100 \%,
	\label{eq:percent}
\end{align}
where $d_1$ and $\hat{d}_1$ are ground truth and  predicted object depth at final pose $z_1$.
%where $d_1$ is ground truth and $\hat{d}_1$ is predicted object depth at the final pose $z_1$.

\subsection{ODMS Test Results}

\setlength{\tabcolsep}{6.75pt}
\begin{table} [t]
	\centering
	\small
	\caption{\textbf{ODMS Test Set Results}	
	}
	\begin{tabular}{| l | c | c | c| c | c | c | c |}
		\hline	\multicolumn{1}{|c|}{} & Object &	$n$	&	\multicolumn{5}{c|}{ Mean Percent Error  \eqref{eq:percent} }  \\	\cline{4-8}
		\multicolumn{1}{|c|}{Config.}	& Depth &  Input  & \multicolumn{1}{c|}{Robot} & \multicolumn{1}{c|}{Driving} & \multicolumn{2}{c|}{Simulated Objects} & All \\ \cline{6-7}
		\multicolumn{1}{|c|}{ID} & Method  &	Masks  &	\multicolumn{1}{c|}{Objects}  & \multicolumn{1}{c|}{Objects}	&	\multicolumn{1}{c|}{Normal} & \multicolumn{1}{c|}{Perturb} & Sets \\	\hline
		\rowcolor{rowgray}	ODN$_\ell$	&	$\mathcal{L}_{\ell}$ \eqref{eq:scaleloss}	&	10	&	19.3	& \bf	30.1	&	8.3	&	18.2	& \bf	19.0	\\	
		ODN$_{\bar{d}}$	&	$\mathcal{L}_{\bar{d}}$  \eqref{eq:normloss} 	&	10	&	18.5	&	30.9	&	8.2	&	18.5	&	19.0	\\	
		\rowcolor{rowgray}	ODN$_d$	&	${\mathcal{L}_{d}}$ \eqref{eq:depthloss}	&	10	& \bf	18.1	&	47.5	& \bf	5.1	& \bf	11.2	&	20.5	\\	
		VOS-DE 	&	\cite{GrFlCo20}	&	10	&	32.6	&	36.0	&	7.9	&	33.6	&	27.5	\\	\hline
		%		VOS-DE \cite{GrFlCo20}	&	LS \eqref{eq:axb}	&	10	&	32.6	&	36.0	&	7.9	&	33.6	&	27.5	\\	\hline
	\end{tabular}
	\label{tab:main}
\end{table}

Object depth estimation results for all four ODMS test sets are provided in Table~\ref{tab:main} for our three ODN configurations and VOS-DE \cite{GrFlCo20}.
We use $n=10$ observations, and ``All Sets'' is an aggregate score across all test sets.
Notably, VOS-DE uses only the largest connected region of each mask to reduce noise.

The relative scale-based ODN$_\ell$ performs best on the Driving set and overall.
We show a few quantitative depth estimation examples for ODN$_\ell$ in Fig.~\ref{fig:val_test} and Fig.~\ref{fig:val_synthia}.
Normalized depth-based ODN$_{\bar{d}}$ comes in second overall, and depth-based ODN$_{d}$ performs best in three categories but worst in driving.
Basically, ODN$_{d}$ gets a performance boost from a camera movement range- and depth-based prior (i.e., $\Delta \mathbf{z}$ and $f_d$ in \eqref{eq:depthloss}) at the cost of applicability to other domains where the scale of camera input and depth will vary.
On the other hand, the generalization of ODN$_{\bar{d}}$ and ODN$_\ell$ from small distances in training to large distances in Driving is highly encouraging.
VOS-DE performs the worst overall, particularly on test sets with segmentation errors or moving objects.
However, VOS-DE does perform well on normal simulated objects, which only have mask discretization errors.

\subsubsection{Results on Changing the Number of Observations}
\label{sec:exp_n}

Object depth estimation results for varied number of observations are provided in Table~\ref{tab:num}.
We repeat training and validation for each new configuration to learn depth estimation with less observations.
As $n$ changes, each test set example uses the same endpoint observations (i.e., $\mathbf{M}_1,\mathbf{M}_n,z_1,z_n$).
However, the $n-2$ intermediate observations are evenly distributed and do change (e.g., $n=2$ has none).
Notably, at $n=2$, VOS-DE is equivalent to \eqref{eq:zobj} and $\mathbf{\bar{z}} \in \mathbb{R}^{n-2}$ \eqref{eq:znorm} gives no input to ODN$_{\bar{d}}$, ODN$_\ell$.

ODN$_\ell$ has the most consistent and best performance for all $n$ settings, aside from a second place to ODN$_{\bar{d}}$ at $n=5$.
ODN$_\ell$ also requires the fewest training iterations for all $n$.
In general, ODN$_{\bar{d}}$ and ODN$_{d}$ performance starts to decrease for $n \leq 3$. 
VOS-DE performance decreases most significantly at $n=2$, having 2.5 times the error of ODN$_\ell$ at $n=2$.
Amazingly, all $n=2$ ODN configurations outperform $n=10$ VOS-DE.
Thus, even with significantly less input data, our learning-based approach outperforms prior work.

\setlength{\tabcolsep}{4pt} 
\begin{table} [t]
	\centering
	\small
	\caption{\textbf{ODMS Test Set Results vs. Number of Observations}	
	}
	\begin{tabular}{| l | c | c | c| c | c | c | c | c | c |}
		\hline
		\multicolumn{1}{|c|}{Config.}	& Depth &  	\multicolumn{4}{c|}{ Overall Mean Percent Error }  & \multicolumn{4}{c|}{ Average Training Iterations}\\	\cline{3-10}
		\multicolumn{1}{|c|}{ID} & Method  &	$n=2$  &	$n=3$ &	$n=5$ &	$n=10$   & $n=2$  &	$n=3$ &	$n=5$ &	$n=10$ \\	\hline
		\rowcolor{rowgray}	ODN$_\ell$	&	$\mathcal{L}_{\ell}$ \eqref{eq:scaleloss}	& \bf	20.4	& \bf	19.9	&	20.0	& \bf	19.0	& \bf	2,590	& \bf	3,460	& \bf	3,060	& \bf	3,138	\\	
		ODN$_{\bar{d}}$	&	$\mathcal{L}_{\bar{d}}$  \eqref{eq:normloss} 	&	22.7	&	20.9	& \bf	19.9	&	19.0	&	3,993	&	4,330	&	3,265	&	3,588	\\	
		\rowcolor{rowgray}	ODN$_d$	&	${\mathcal{L}_{d}}$ \eqref{eq:depthloss}	&	21.6	&	21.2	&	20.5	&	20.5	&	4,138	&	4,378	&	4,725	&	3,300	\\	
		VOS-DE	&	 \cite{GrFlCo20}	&	50.3	&	29.7	&	27.6	&	27.5	&	N/A	&	N/A	&	N/A	&	N/A	\\	\hline
		%VOS-DE \cite{GrFlCo20}	&	LS \eqref{eq:axb}	&	50.3	&	29.7	&	27.6	&	27.5	&	N/A	&	N/A	&	N/A	&	N/A	\\	\hline
	\end{tabular}
	\label{tab:num}
\end{table}

\setlength{\tabcolsep}{4.7pt} 
\begin{table} [t]
	\centering
	\small
	\caption{\textbf{Test Results with Perturb Training Data and Radial Input Image}	
		%		{\color{note} All configurations use $n=10$ observations.}
	}
	\begin{tabular}{| l | c | c | c | c| c | c | c | c |}
		\hline	\multicolumn{1}{|c|}{} & Object &  Radial& Type of &	\multicolumn{5}{c|}{ Mean Percent Error  \eqref{eq:percent} }  \\	\cline{5-9}
		\multicolumn{1}{|c|}{Config.}	& Depth &  Input & Training &  \multicolumn{1}{c|}{} & \multicolumn{1}{c|}{} & \multicolumn{2}{c|}{Simulated} & All \\ \cline{7-8}
		\multicolumn{1}{|c|}{ID} & Method  &  Image & Data &	\multicolumn{1}{c|}{Robot}  & \multicolumn{1}{c|}{Driving}	&	\multicolumn{1}{c|}{Normal} & \multicolumn{1}{c|}{Perturb} & Sets \\	\hline 
		\hline \multicolumn{9}{|c|}{Perturb Training Data}\\ \hline																		
		\rowcolor{rowgray}	ODN$_{\ell p}$	&	$\mathcal{L}_{\ell}$ \eqref{eq:scaleloss}	&	No	&	Perturb	&	22.2	& \bf	29.0	&	11.1	&	13.0	&	18.8	\\	
		ODN$_{\bar{d} p}$	&	$\mathcal{L}_{\bar{d}}$ \eqref{eq:normloss} 	&	No	&	Perturb	&	25.8	&	31.4	&	11.1	&	13.2	&	20.4	\\	
		\rowcolor{rowgray}	ODN$_{d p}$	&	${\mathcal{L}_{d}}$ \eqref{eq:depthloss}	&	No	&	Perturb	&	20.1	&	60.9	&	7.3	& \bf	8.2	&	24.1	\\	\hline
		\hline \multicolumn{9}{|c|}{Radial Input Image}\\ \hline																		
		\rowcolor{rowgray}	ODN$_{\ell r}$	&	$\mathcal{L}_{\ell}$ \eqref{eq:scaleloss}	&	Yes	&	Normal	& \bf	13.1	&	31.7	&	8.6	&	17.9	& \bf	17.8	\\	
		ODN$_{\bar{d} r}$	&	$\mathcal{L}_{\bar{d}}$  \eqref{eq:normloss} 	&	Yes	&	Normal	&	15.2	&	30.9	&	8.4	&	18.5	&	18.3	\\	
		\rowcolor{rowgray}	ODN$_{d r}$	&	${\mathcal{L}_{d}}$ \eqref{eq:depthloss}	&	Yes	&	Normal	&	13.4	&	48.6	& \bf	5.6	&	11.2	&	19.7	\\	\hline
		%\hline \multicolumn{9}{|c|}{Standard Configuration}\\ \hline																		
		%\rowcolor{rowgray}	ODN$_\ell$	&	$\mathcal{L}_{\ell}$ \eqref{eq:scaleloss}	&	No	&	Normal	&	19.3	&	30.1	&	8.3	&	18.2	&	19.0	\\	
		%ODN$_{\bar{Z}}$	&	$\mathbf{\mathcal{L}_{\bar{Z}}}$ \eqref{eq:normloss} 	&	No	&	Normal	&	18.5	&	30.9	&	8.2	&	18.5	&	19.0	\\	
		%\rowcolor{rowgray}	ODN$_Z$	&	$\mathbf{\mathcal{L}_{Z}}$ \eqref{eq:depthloss}	&	No	&	Normal	&	18.1	&	47.5	& \bf	5.1	&	11.2	&	20.5	\\	\hline
	\end{tabular}
	\label{tab:rad}
\end{table}

\subsubsection{Results with Perturbation Training Data}
We train each $n=10$ ODN on continuously-generated perturbation data \eqref{eq:perturb} from Section~\ref{sec:sim_test}.
As shown in Table~\ref{tab:rad}, this improves performance for each ODN on the Perturb test set, and demonstrates that we can learn robust depth estimation for specific errors.
The perturbed ODN$_\ell$ configuration, ODN$_{\ell p}$, improves performance overall and has the best Driving result of any method.

\subsubsection{Results with Radial Input Image}
\label{sec:radial}
For our final ODMS results in Table~\ref{tab:rad}, we train each $n=10$ ODN with an added input radial image for convolution.
Pixel values $\in [0,1]$ are scaled radially from 1 at the center to 0 at each corner (see Fig.~\ref{fig:network}).
This serves a similar purpose to coordinate convolution \cite{LiEtAl18} but simply focuses on how centered segmentation mask regions are.
This improves overall performance for each ODN, particularly on the Robot test, where objects are generally centered for grasping and peripheral segmentation errors can be ignored.
Notably, ODN$_{\ell r}$ has the best Robot and overall result of any method.

\subsection{Robot Object Depth Estimation and Grasping Experiments}
\label{sec:robot_exp}

As a live robotics demonstration, we use ODN$_{\ell r}$ to locate objects for grasping.
Experiments start with HSR's grasp camera approaching an object while generating segmentation masks at 1~\textrm{cm} increments using pre-trained OSVOS.
Once ten masks are available, ODN$_{\ell r}$ starts predicting depth as HSR continues approaching and generating masks. 
%ODN$_{\ell r}$'s prediction speed is negligible compared to HSR's data-collection speed, so we use the median depth estimate of multiple permutations of collected data to improve robustness against segmentation errors.
Because ODN$_{\ell r}$'s prediction speed is negligible compared to HSR's data-collection speed, we use the median depth estimate of multiple permutations of collected data to improve robustness against segmentation errors.
Once ODN$_{\ell r}$ estimates the object to be within 20~\textrm{cm} of grasping, HSR stops collecting data and grasps the object at that depth.
Using this active depth estimation process, we are able to successfully locate and grasp consecutive objects at varied heights in a variety of settings, including placing laundry in a basket and clearing garbage off a table (see Fig.~\ref{fig:front_vos}).
We show these robot experiments in our Supplementary Video at: \url{https://youtu.be/c90Fg_whjpI}. %Supplementary Video).

\section{Conclusions}

We introduce the \textbf{O}bject \textbf{D}epth via \textbf{M}otion and \textbf{S}egmentation (ODMS) dataset, which continuously generates synthetic training data with random camera motion, objects, and even perturbations.
Using the ODMS dataset, we train the first deep network to estimate object depth from motion and segmentation, leading to as much as a 59\% reduction in error over previous work.
By using ODMS's simple binary mask- and distance-based input, our network's performance transfers across sim-to-real and diverse application domains, as demonstrated by our results on the robotics-, driving-, and simulation-based ODMS test sets.
Finally, we use our network to perform object depth estimation in real-time robot grasping experiments, demonstrating how our segmentation-based approach to depth estimation is a viable tool for real-world applications requiring 3D perception from a single RGB camera.
%In future work, we will expand the ODMS dataset to include new types of training data with perturbations that better represent segmentation errors in robotics and driving. 

\subsection*{Acknowledgements}
We thank Madan Ravi Ganesh, Parker Koch, and Luowei Zhou for various discussions throughout this work. Toyota Research Institute (``TRI") provided funds to assist the authors with their research but this article solely reflects the opinions and conclusions of its authors and not TRI or any other Toyota entity.

\section*{Appendix}

\subsection*{Least-squares Solution for Object Depth}

In previous work \cite{GrFlCo20}, we propose a least-squares object depth solution (VOS-DE) that uses more than two observations to add robustness for camera position and segmentation errors. 
We include this solution here for reference.
The VOS-DE formulation derives an alternative form of \eqref{eq:cobj} as
\begin{align}
	z_{\text{object}} \sqrt{a_i}   + c = z_i  \sqrt{a_i},
\end{align}
which over $n$ observations in $\mathbf{A}\mathbf{x}=\mathbf{b}$ form yields
\begin{align}
	\begin{bmatrix}
		\sqrt{a_1} & 1 \\ \sqrt{a_2} & 1 \\ \vdots & \vdots \\ \sqrt{a_n} & 1
	\end{bmatrix}
	\begin{bmatrix}
		\hat{z}_{\text{object}} \\ \hat{c}\\ 
	\end{bmatrix}
	=
	\begin{bmatrix}
		z_{1} \sqrt{a_1} \\ z_{2} \sqrt{a_2} \\ \vdots \\ z_{n} \sqrt{a_n}
	\end{bmatrix}.
	\label{eq:axb}
\end{align}
Solving \eqref{eq:axb} for $\hat{z}_{\text{object}}$ does provide a more robust depth estimate than the two-observation solution \eqref{eq:zobj}. 
However, our learning-based approach from Section~\ref{sec:learn} outperforms both analytic solutions in experiments.

\subsection*{ODMS Random Object Mask Examples}

\begin{figure} 
	\centering
	\includegraphics[width=0.975\textwidth]{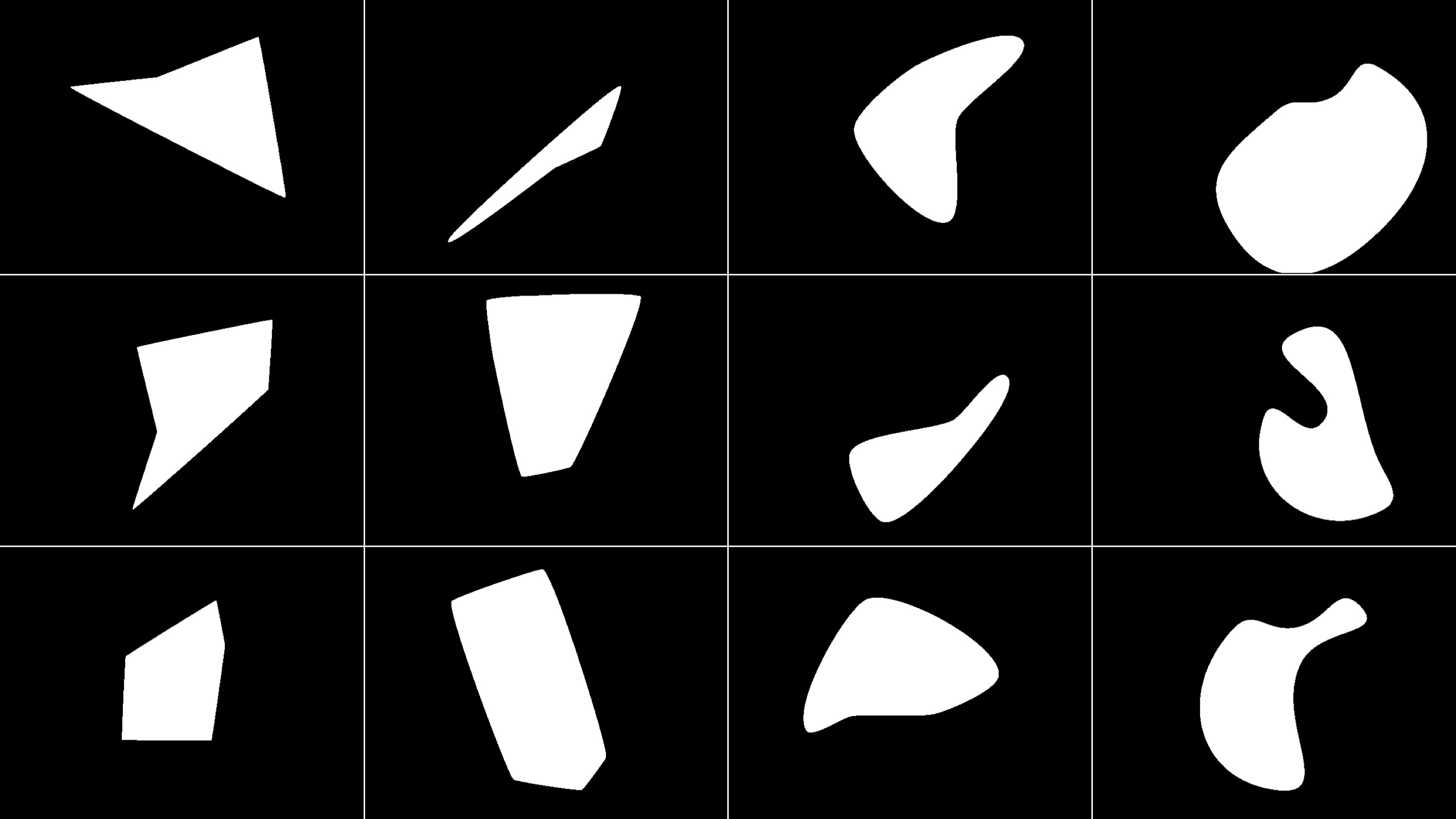}
	\caption{ \textbf{ODMS Random Object Mask Examples.}
		All examples use $s_\mathbf{p}=400$, $n_\mathbf{p}=5$, and $\ell=\ell_\text{min}=1$.  
		$r_B$ values are 0.01, 0.05, 0.2, and 0.5 (\textit{from left to right}) and $\rho_B$ values are 0.01, 0.05, and 0.2 (\textit{from top to bottom}).
		Each generated object is unique
	}
	\label{fig:synth_obj}
\end{figure}

We provide a few random object mask examples using ODMS's data-generation framework from Section~\ref{sec:gen_masks}.
These synthetic object examples are shown in Fig.~\ref{fig:synth_obj} and demonstrate the B\'ezier curve behaviors associated with changing parameters $r_B$ and $\rho_B$.

\subsection*{ODMS Validation Results}
%The complete ODMS validation and test set results for all ODN and VOS-DE configurations are provided in Table~\ref{tab:all}.

As mentioned in Section~\ref{sec:results}, the number of network training iterations is determined by the best validation performance, which we check at every ten training iterations.
In Table~\ref{tab:all}, we provide the ODMS validation results and corresponding number of training iterations for all configurations.
In general, the relative performance of each configuration is consistent between the ODMS validation and test sets.

\setlength{\tabcolsep}{1.75pt} 
\setlength{\tabcolsep}{4pt} 
\begin{table}
	\centering
	\scriptsize
	\caption{\textbf{Complete ODMS Validation and Test Set Results}	
	}
	\begin{tabular}{| l | c | c | c | c | c | c | c | c |}
		\hline
		\multicolumn{1}{|c|}{Config.}	 & 	\multicolumn{4}{c|}{ Mean Percent Error (Validation/Test) }  & \multicolumn{4}{c|}{ Training Iterations}\\	\cline{2-9}
		\multicolumn{1}{|c|}{ID}   &	Robot  &	Driving &	Normal  &	Perturb   & Robot  &	Driving &	Normal  &	Perturb \\	\hline
		\hline \multicolumn{9}{|c|}{Standard Configuration}																		\\	\hline
		\rowcolor{rowgray}	ODN$_\ell$	&	21.6/19.3	&	29.4/30.1	&	8.2/8.3	&	18.4/18.2	&	2390	&	1920	&	3370	&	4870	\\	
		ODN$_{\bar{d}}$ 	&	19.6/18.5	&	32.0/30.9	&	7.9/8.2	&	18.4/18.5	&	4140	&	2990	&	3690	&	3530	\\	
		\rowcolor{rowgray}	ODN$_d$ 	&	19.9/18.1	&	48.1/47.5	& \bf	4.9/5.1	&	11.5/11.2	&	2380	&	1650	&	4740	&	4430	\\	
		\scriptsize VOS-DE 	&	27.4/32.6	&	35.9/36.0	&	7.9/7.9	&	34.1/33.6	&	N/A	&	N/A	&	N/A	&	N/A	\\	\hline
		\hline \multicolumn{9}{|c|}{$n=5$ Observations}																		\\	\hline
		\rowcolor{rowgray}	ODN$_{\ell}$ 	&	23.4/20.5	&	31.5/30.5	&	8.4/8.6	&	20.2/20.4	&	1000	&	1850	&	4870	&	4520	\\	
		ODN$_{\bar{d}}$ 	&	22.8/19.5	&	34.2/31.1	&	8.4/8.4	&	20.5/20.6	&	1510	&	3450	&	3770	&	4330	\\	
		\rowcolor{rowgray}	ODN$_{d}$	&	21.0/19.4	&	44.6/44.2	&	5.4/5.5	&	13.4/12.9	&	4690	&	4260	&	4980	&	4970	\\	
		\scriptsize VOS-DE	&	29.5/35.1	&	34.8/34.6	&	7.8/7.9	&	32.8/32.6	&	N/A	&	N/A	&	N/A	&	N/A	\\	\hline
		\hline \multicolumn{9}{|c|}{$n=3$ Observations}																		\\	\hline
		\rowcolor{rowgray}	ODN$_{\ell}$	&	20.3/18.6	&	31.8/31.1	&	8.4/8.4	&	21.9/21.6	&	1820	&	2750	&	4890	&	4380	\\	
		ODN$_{\bar{d}}$ 	&	19.9/20.6	&	34.7/33.1	&	8.4/8.4	&	21.6/21.5	&	4130	&	4320	&	4620	&	4250	\\	
		\rowcolor{rowgray}	ODN$_{d}$ 	&	24.0/21.8	&	45.1/44.5	&	5.4/5.6	&	13.8/12.9	&	4800	&	3040	&	4990	&	4680	\\	
		\scriptsize VOS-DE 	&	33.7/41.2	&	45.2/34.0	&	8.0/8.1	&	37.0/35.7	&	N/A	&	N/A	&	N/A	&	N/A	\\	\hline
		\hline \multicolumn{9}{|c|}{$n=2$ Observations}																		\\	\hline
		\rowcolor{rowgray}	ODN$_{\ell}$	&	21.3/19.2	&	30.4/31.4	&	8.7/8.9	&	22.0/22.0	&	1140	&	1010	&	3910	&	4300	\\	
		ODN$_{\bar{d}}$ 	&	29.1/24.2	&	39.6/35.9	&	8.6/8.9	&	21.8/21.8	&	3410	&	4570	&	3370	&	4620	\\	
		\rowcolor{rowgray}	ODN$_{d}$	&	23.3/21.1	&	45.3/44.8	&	5.8/6.0	&	14.9/14.4	&	2850	&	4120	&	4610	&	4970	\\	
		\scriptsize VOS-DE 	&	95.8/65.5	&	55.0/41.1	&	8.2/8.3	&	90.6/86.2	&	N/A	&	N/A	&	N/A	&	N/A	\\	\hline
		\hline \multicolumn{9}{|c|}{Perturb Training Data}																		\\	\hline
		\rowcolor{rowgray}	ODN$_{\ell p}$	&	21.4/22.2	& \bf	28.6/29.0	&	10.7/11.1	&	12.8/13.0	&	100	&	140	&	5000	&	5000	\\	
		ODN$_{\bar{d} p}$	&	25.6/25.8	&	31.4/31.4	&	11.0/11.1	&	13.1/13.2	&	420	&	2760	&	2730	&	4270	\\	
		\rowcolor{rowgray}	ODN$_{d p}$	&	20.5/20.1	&	59.4/60.9	&	7.0/7.3	& \bf	8.1/8.2	&	50	&	330	&	4860	&	4780	\\	\hline
		\hline \multicolumn{9}{|c|}{Radial Input Image}																		\\	\hline
		\rowcolor{rowgray}	ODN$_{\ell r}$	& \bf	13.8/13.1	&	31.6/31.7	&	8.4/8.6	&	18.2/17.9	&	1710	&	870	&	4940	&	3940	\\	
		ODN$_{\bar{d} r}$	&	16.6/15.2	&	30.7/30.9	&	8.3/8.4	&	18.6/18.5	&	2010	&	4200	&	4990	&	4440	\\	
		\rowcolor{rowgray}	ODN$_{d r}$	&	14.1/13.4	&	49.0/48.6	&	5.5/5.6	&	11.7/11.2	&	2210	&	460	&	4870	&	4710	\\	\hline
	\end{tabular}
	\label{tab:all}
\end{table}

\subsection*{ODMS Absolute Error Results}

In Table~\ref{tab:abs}, we provide ODMS test results for the mean absolute error, which is calculated for each example as
\begin{align}
	\text{Absolute Error} = \left| d_1 - \hat{d}_1 \right|,
	\label{eq:abs}
\end{align}
where $d_1$ and $\hat{d}_1$ are ground truth and  predicted object depth at final pose $z_1$.
Notably, our motivation to use percent error \eqref{eq:percent} in the paper is to provide a consistent comparison across domains with markedly different object depth distances.
For example, the 6 \textrm{cm} absolute error from Fig.~\ref{fig:val_synthia} is much better for the driving domain than it would be for robot grasping.

\setlength{\tabcolsep}{1.75pt} 
\setlength{\tabcolsep}{4pt} 
\begin{table}
	\centering
	\scriptsize
	\caption{\textbf{Complete ODMS Validation and Test Set Results (Absolute Error)}	
	}
	\begin{tabular}{| l | c | c | c | c | c | c | c | c |}
		\hline
		\multicolumn{1}{|c|}{}	 & 	\multicolumn{4}{c|}{ Mean Absolute Error (Validation/Test) }  & \multicolumn{4}{c|}{ Training Iterations}\\	\cline{2-9}
		\multicolumn{1}{|c|}{Config.}   &	Robot  &	Driving &	Normal  &	Perturb   &   &	 &	  &	 \\	
		\multicolumn{1}{|c|}{ID}   &	\textrm{(cm)}  &	\textrm{(m)}  &	\textrm{(cm)}   &	\textrm{(cm)}    & Robot  &	Driving &	Normal  &	Perturb \\	\hline
		\hline \multicolumn{9}{|c|}{Standard Configuration}																		\\	\hline
		\rowcolor{rowgray}	ODN$_\ell$	&	7.2/6.6	&	3.8/4.3	&	3.4/3.4	&	7.3/7.2	&	2390	&	1920	&	3370	&	4870	\\	
		ODN$_{\bar{d}}$ 	&	6.4/6.0	&	4.1/4.4	&	3.1/3.1	&	7.4/7.3	&	4140	&	2990	&	3690	&	3530	\\	
		\rowcolor{rowgray}	ODN$_d$ 	&	6.8/6.3	&	7.1/7.8	& \bf	1.8/1.8	&	3.9/3.7	&	2380	&	1650	&	4740	&	4430	\\	
		\scriptsize VOS-DE 	&	8.8/10.0	&	5.0/5.4	&	2.8/2.8	&	15.3/14.9	&	N/A	&	N/A	&	N/A	&	N/A	\\	\hline
		\hline \multicolumn{9}{|c|}{$n=5$ Observations}																		\\	\hline
		\rowcolor{rowgray}	ODN$_{\ell 5}$ 	&	7.8/7.0	&	3.9/4.3	&	3.4/3.5	&	8.2/8.1	&	1000	&	1850	&	4870	&	4520	\\	
		ODN$_{\bar{d}5}$ 	&	7.0/6.1	&	4.4/4.6	&	3.3/3.3	&	8.1/8.0	&	1510	&	3450	&	3770	&	4330	\\	
		\rowcolor{rowgray}	ODN$_{d 5}$	&	7.3/6.8	&	6.5/7.2	&	1.9/2.0	&	4.9/4.7	&	4690	&	4260	&	4980	&	4970	\\	
		\scriptsize VOS-DE$_5$	&	9.6/10.8	&	5.0/5.2	&	2.9/2.9	&	14.3/14.2	&	N/A	&	N/A	&	N/A	&	N/A	\\	\hline
		\hline \multicolumn{9}{|c|}{$n=3$ Observations}																		\\	\hline
		\rowcolor{rowgray}	ODN$_{\ell 3}$	&	6.8/6.3	&	4.1/4.5	&	3.4/3.4	&	8.8/8.6	&	1820	&	2750	&	4890	&	4380	\\	
		ODN$_{\bar{d}3}$ 	&	6.8/7.0	&	4.4/4.7	&	3.3/3.3	&	8.6/8.4	&	4130	&	4320	&	4620	&	4250	\\	
		\rowcolor{rowgray}	ODN$_{d 3}$ 	&	7.9/7.3	&	6.6/7.3	&	1.9/1.9	&	4.8/4.4	&	4800	&	3040	&	4990	&	4680	\\	
		\scriptsize VOS-DE$_3$ 	&	11.2/12.6	&	6.3/5.0	&	2.9/2.9	&	15.7/15.3	&	N/A	&	N/A	&	N/A	&	N/A	\\	\hline
		\hline \multicolumn{9}{|c|}{$n=2$ Observations}																		\\	\hline
		\rowcolor{rowgray}	ODN$_{\ell 2}$	&	7.0/6.4	&	3.7/4.3	&	3.5/3.6	&	8.5/8.4	&	1140	&	1010	&	3910	&	4300	\\	
		ODN$_{\bar{d}2}$ 	&	9.2/7.8	&	4.8/5.0	&	3.5/3.5	&	8.6/8.4	&	3410	&	4570	&	3370	&	4620	\\	
		\rowcolor{rowgray}	ODN$_{d 2}$	&	8.0/7.2	&	6.8/7.5	&	2.0/2.1	&	5.4/5.1	&	2850	&	4120	&	4610	&	4970	\\	
		\scriptsize VOS-DE$_2$ 	&	36.2/21.9	&	8.5/6.7	&	3.0/3.0	&	42.1/39.7	&	N/A	&	N/A	&	N/A	&	N/A	\\	\hline
		\hline \multicolumn{9}{|c|}{Perturb Training Data}																		\\	\hline
		\rowcolor{rowgray}	ODN$_{\ell p}$	&	7.0/6.9	& \bf	3.5/4.1	&	4.3/4.5	&	5.2/5.2	&	100	&	140	&	5000	&	5000	\\	
		ODN$_{\bar{d} p}$	&	8.4/8.5	&	4.0/4.4	&	4.4/4.4	&	5.2/5.1	&	420	&	2760	&	2730	&	4270	\\	
		\rowcolor{rowgray}	ODN$_{d p}$	&	6.7/5.8	&	8.9/9.9	&	2.4/2.5	&	\bf 2.8/2.8	&	50	&	330	&	4860	&	4780	\\	\hline
		\hline \multicolumn{9}{|c|}{Radial Input Image}																		\\	\hline
		\rowcolor{rowgray}	ODN$_{\ell r}$	& \bf	4.4/4.3	&	4.0/4.5	&	3.5/3.5	&	7.4/7.2	&	1710	&	870	&	4940	&	3940	\\	
		ODN$_{\bar{d} r}$	&	5.6/5.0	&	3.8/4.3	&	3.3/3.4	&	7.5/7.4	&	2010	&	4200	&	4990	&	4440	\\	
		\rowcolor{rowgray}	ODN$_{d r}$	&	4.4/4.4	&	7.2/8.0	&	1.9/1.9	&	4.3/4.0	&	2210	&	460	&	4870	&	4710	\\	\hline
	\end{tabular}
	\label{tab:abs}
\end{table}

\subsection*{ODMS Robot Test Set Segmentation Examples}

\begin{figure}
	\centering
	\includegraphics[width=0.975\textwidth]{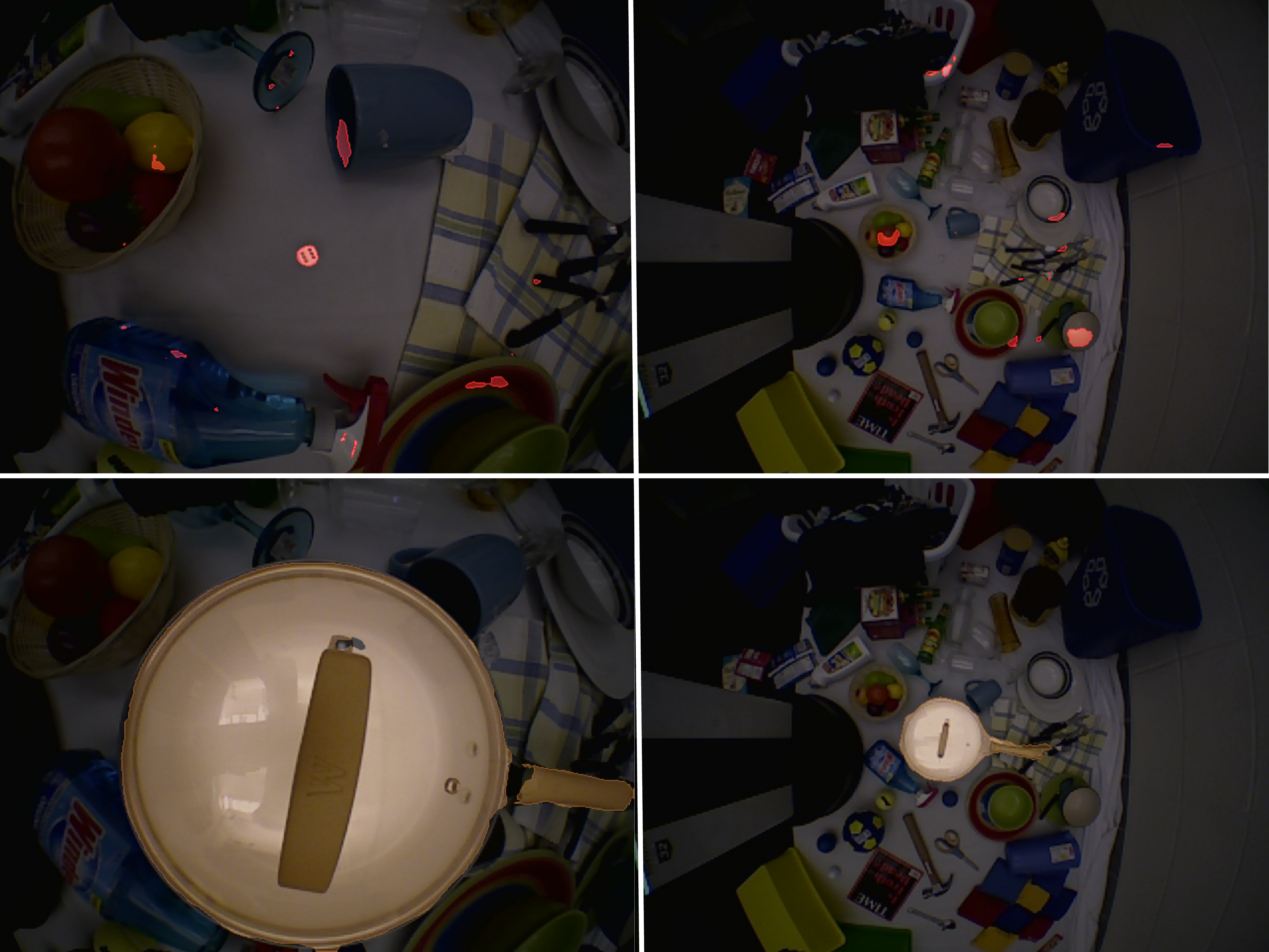}
	\caption{ \textbf{ODMS Robot Test Set Segmentation Examples.}
		The small die segmentation (\textit{top}) has fragments of other objects in the closest view (\textit{left}) and completely misses the die in the farthest view (\textit{right}).
		On the other hand, the larger pan segmentation (\textit{bottom}) misses parts of the handle that are out of the image in the closest view (\textit{left}) but is fairly accurate in the farthest view (\textit{right})
	}
	\label{fig:test_obj}
\end{figure}

For the ODMS Robot test set, we intentionally choose challenging objects, spanning from a single die to the 470~\textrm{mm} long pan.
Not surprising, segmenting diverse objects presents varied challenges.
To illustrate this point, in Fig.~\ref{fig:test_obj} we show the closest and farthest Robot test set segmentations for the die and pan.

\bibliographystyle{splncs04}
\bibliography{GrCoECCV20}

\begin{thebibliography}{10}
\providecommand{\url}[1]{\texttt{#1}}
\providecommand{\urlprefix}{URL }
\providecommand{\doi}[1]{https://doi.org/#1}

\bibitem{CINM}
Bao, L., Wu, B., Liu, W.: {CNN} in {MRF:} video object segmentation via
  inference in {A} cnn-based higher-order spatio-temporal {MRF}. In: IEEE
  Conference on Computer Vision and Pattern Recognition (CVPR) (2018)

\bibitem{OSVOS}
Caelles, S., Maninis, K.K., Pont-Tuset, J., Leal-Taix\'e, L., Cremers, D., {Van
  Gool}, L.: One-shot video object segmentation. In: IEEE Conference on
  Computer Vision and Pattern Recognition (CVPR) (2017)

\bibitem{DAVIS2018}
Caelles, S., Montes, A., Maninis, K., Chen, Y., Gool, L.V., Perazzi, F.,
  Pont{-}Tuset, J.: The 2018 {DAVIS} challenge on video object segmentation.
  CoRR  \textbf{abs/1803.00557} (2018)

\bibitem{YCB}
{Calli}, B., {Walsman}, A., {Singh}, A., {Srinivasa}, S., {Abbeel}, P.,
  {Dollar}, A.M.: Benchmarking in manipulation research: Using the
  yale-cmu-berkeley object and model set. IEEE Robotics Automation Magazine
  \textbf{22}(3),  36--52 (2015)

\bibitem{ChChChACMM2012}
Chen, D.J., Chen, H.T., Chang, L.W.: Video object cosegmentation. In: ACM
  International Conference on Multimedia (2012)

\bibitem{PML}
Chen, Y., Pont-Tuset, J., Montes, A., {Van Gool}, L.: Blazingly fast video
  object segmentation with pixel-wise metric learning. In: Computer Vision and
  Pattern Recognition (CVPR) (2018)

\bibitem{City}
Cordts, M., Omran, M., Ramos, S., Rehfeld, T., Enzweiler, M., Benenson, R.,
  Franke, U., Roth, S., Schiele, B.: The cityscapes dataset for semantic urban
  scene understanding. In: Proc. of the IEEE Conference on Computer Vision and
  Pattern Recognition (CVPR) (2016)

\bibitem{ImageNet}
Deng, J., Dong, W., Socher, R., Li, L.J., Li, K., Fei-Fei, L.: Imagenet: A
  large-scale hierarchical image database. In: IEEE Conference on Computer
  Vision and Pattern Recognition (CVPR) (2009)

\bibitem{NLC}
Faktor, A., Irani, M.: Video segmentation by non-local consensus voting. In:
  British Machine Vision Conference (BMVC) (2014)

\bibitem{FeLa19}
{Ferguson}, M., {Law}, K.: A 2d-3d object detection system for updating
  building information models with mobile robots. In: 2019 IEEE Winter
  Conference on Applications of Computer Vision (WACV) (2019)

\bibitem{FlCoGr19}
Florence, V., Corso, J.J., Griffin, B.: Robot-supervised learning for object segmentation. In: The IEEE International Conference on Robotics and Automation
  (ICRA) (2020)

\bibitem{GaEtAl20}
{Gan}, L., {Zhang}, R., {Grizzle}, J.W., {Eustice}, R.M., {Ghaffari}, M.:
  Bayesian spatial kernel smoothing for scalable dense semantic mapping. IEEE
  Robotics and Automation Letters (RA-L)  \textbf{5}(2) (2020)

\bibitem{KITTI}
{Geiger}, A., {Lenz}, P., {Urtasun}, R.: Are we ready for autonomous driving?
  the kitti vision benchmark suite. In: 2012 IEEE Conference on Computer Vision
  and Pattern Recognition (CVPR) (2012)

\bibitem{GrFlCo20}
Griffin, B., Florence, V., Corso, J.J.: Video object segmentation-based visual
  servo control and object depth estimation on a mobile robot. In: IEEE Winter
  Conference on Applications of Computer Vision (WACV) (2020)

\bibitem{GrCo19}
Griffin, B.A., Corso, J.J.: Bubblenets: Learning to select the guidance frame
  in video object segmentation by deep sorting frames. In: The IEEE Conference
  on Computer Vision and Pattern Recognition (CVPR) (2019)

\bibitem{GrCoWACV2019}
Griffin, B.A., Corso, J.J.: Tukey-inspired video object segmentation. In: IEEE
  Winter Conference on Applications of Computer Vision (WACV) (2019)

\bibitem{maskrcnn}
He, K., Gkioxari, G., Dollar, P., Girshick, R.: Mask r-cnn. In: The IEEE
  International Conference on Computer Vision (ICCV) (2017)

\bibitem{HeEtAl16}
He, K., Zhang, X., Ren, S., Sun, J.: Deep residual learning for image
  recognition. In: The IEEE Conference on Computer Vision and Pattern
  Recognition (CVPR) (2016)

\bibitem{It51}
Ittelson, W.H.: Size as a cue to distance: Radial motion. The American Journal
  of Psychology  \textbf{64}(2),  188--202 (1951)

\bibitem{KaGaBa19}
{Kasten}, Y., {Galun}, M., {Basri}, R.: Resultant based incremental recovery of
  camera pose from pairwise matches. In: IEEE Winter Conference on Applications
  of Computer Vision (WACV) (2019)

\bibitem{KhEtAl19}
Khan, S.H., Guo, Y., Hayat, M., Barnes, N.: Unsupervised primitive discovery
  for improved 3d generative modeling. In: The IEEE Conference on Computer
  Vision and Pattern Recognition (CVPR) (2019)

\bibitem{adam14}
Kingma, D.P., Ba, J.: Adam: {A} method for stochastic optimization. In:
  International Conference on Learning Representations (ICLR) (2014)

\bibitem{SuSh19}
Kumawat, S., Raman, S.: Lp-3dcnn: Unveiling local phase in 3d convolutional
  neural networks. In: The IEEE Conference on Computer Vision and Pattern
  Recognition (CVPR) (2019)

\bibitem{KEY}
Lee, Y.J., Kim, J., Grauman, K.: Key-segments for video object segmentation.
  In: IEEE International Conference on Computer Vision (ICCV) (2011)

\bibitem{SegTrackv2}
Li, F., Kim, T., Humayun, A., Tsai, D., Rehg, J.M.: Video segmentation by
  tracking many figure-ground segments. In: The IEEE International Conference
  on Computer Vision (ICCV) (2013)

\bibitem{LiEtAl19}
Li, Z., Dekel, T., Cole, F., Tucker, R., Snavely, N., Liu, C., Freeman, W.T.:
  Learning the depths of moving people by watching frozen people. In: The IEEE
  Conference on Computer Vision and Pattern Recognition (CVPR) (2019)

\bibitem{MSCOCO}
Lin, T.Y., Maire, M., Belongie, S., Hays, J., Perona, P., Ramanan, D.,
  Doll{\'a}r, P., Zitnick, C.L.: Microsoft coco: Common objects in context. In:
  Computer Vision -- (ECCV). pp. 740--755. Springer International Publishing,
  Cham (2014)

\bibitem{LiHeCVPR2015}
Liu, B., He, X.: Multiclass semantic video segmentation with object-level
  active inference. In: IEEE Conference on Computer Vision and Pattern
  Recognition (CVPR) (2015)

\bibitem{LiEtAl18}
Liu, R., Lehman, J., Molino, P., Petroski~Such, F., Frank, E., Sergeev, A.,
  Yosinski, J.: An intriguing failing of convolutional neural networks and the
  coordconv solution. In: Advances in Neural Information Processing Systems 31
  (NIPS)

\bibitem{LiuEtAl19}
Liu, Y., Fan, B., Xiang, S., Pan, C.: Relation-shape convolutional neural
  network for point cloud analysis. In: The IEEE Conference on Computer Vision
  and Pattern Recognition (CVPR) (2019)

\bibitem{LuXuCoCVPR2015}
Lu, J., Xu, R., Corso, J.J.: Human action segmentation with hierarchical
  supervoxel consistency. In: IEEE Conference on Computer Vision and Pattern
  Recognition (CVPR) (2015)

\bibitem{PREMVOS}
Luiten, J., Voigtlaender, P., Leibe, B.: Premvos: Proposal-generation,
  refinement and merging for video object segmentation. In: Asian Conference on
  Computer Vision (ACCV) (2018)

\bibitem{OSVOS-S}
Maninis, K., Caelles, S., Chen, Y., Pont-Tuset, J., Leal-Taixé, L., Cremers,
  D., Gool, L.V.: Video object segmentation without temporal information. IEEE
  Transactions on Pattern Analysis and Machine Intelligence  (2018)

\bibitem{MuTa17}
{Mur-Artal}, R., {Tardós}, J.D.: Orb-slam2: An open-source slam system for
  monocular, stereo, and rgb-d cameras. IEEE Transactions on Robotics (T-RO)
  (2017)

\bibitem{RGMP}
Oh, S.W., Lee, J.Y., Sunkavalli, K., Kim, S.J.: Fast video object segmentation
  by reference-guided mask propagation. In: IEEE Conference on Computer Vision
  and Pattern Recognition (CVPR) (2018)

\bibitem{OnReVeECCV2014}
Oneata, D., Revaud, J., Verbeek, J., Schmid, C.: Spatio-temporal object
  detection proposals. In: European Conference on Computer Vision (ECCV) (2014)

\bibitem{FST}
Papazoglou, A., Ferrari, V.: Fast object segmentation in unconstrained video.
  In: Proceedings of the IEEE International Conference on Computer Vision
  (ICCV) (2013)

\bibitem{DAVIS}
Perazzi, F., Pont-Tuset, J., McWilliams, B., Van~Gool, L., Gross, M.,
  Sorkine-Hornung, A.: A benchmark dataset and evaluation methodology for video
  object segmentation. In: IEEE Conference on Computer Vision and Pattern
  Recognition (CVPR) (2016)

\bibitem{DAVIS17}
Pont{-}Tuset, J., Perazzi, F., Caelles, S., Arbelaez, P., Sorkine{-}Hornung,
  A., Gool, L.V.: The 2017 {DAVIS} challenge on video object segmentation. CoRR
   \textbf{abs/1704.00675} (2017)

\bibitem{pointnet}
Qi, C.R., Su, H., Mo, K., Guibas, L.J.: Pointnet: Deep learning on point sets
  for 3d classification and segmentation. In: The IEEE Conference on Computer
  Vision and Pattern Recognition (CVPR) (2017)

\bibitem{SYNTHIA}
Ros, G., Sellart, L., Materzynska, J., Vazquez, D., Lopez, A.M.: The synthia
  dataset: A large collection of synthetic images for semantic segmentation of
  urban scenes. In: The IEEE Conference on Computer Vision and Pattern
  Recognition (CVPR) (2016)

\bibitem{ScFr16}
Schonberger, J.L., Frahm, J.M.: Structure-from-motion revisited. In: The IEEE
  Conference on Computer Vision and Pattern Recognition (CVPR) (2016)

\bibitem{SoIdShICCV2015}
Soomro, K., Idrees, H., Shah, M.: Action localization in videos through context
  walk. In: IEEE International Conference on Computer Vision (ICCV) (2015)

\bibitem{SoIdShCVPR2016}
Soomro, K., Idrees, H., Shah, M.: Predicting the where and what of actors and
  actions through online action localization. In: IEEE Conference on Computer
  Vision and Pattern Recognition (CVPR) (2016)

\bibitem{dropout14}
Srivastava, N., Hinton, G., Krizhevsky, A., Sutskever, I., Salakhutdinov, R.:
  Dropout: A simple way to prevent neural networks from overfitting. Journal of
  Machine Learning Research  \textbf{15},  1929--1958 (2014)

\bibitem{SwGo86}
Swanston, M.T., Gogel, W.C.: Perceived size and motion in depth from optical
  expansion. Perception \& Psychophysics  \textbf{39},  309--326 (1986)

\bibitem{TaSuYaCVPR2013}
Tang, K., Sukthankar, R., Yagnik, J., Fei-Fei, L.: Discriminative segment
  annotation in weakly labeled video. In: IEEE Conference on Computer Vision
  and Pattern Recognition (CVPR) (2013)

\bibitem{TiLaIJCV2012}
Tighe, J., Lazebnik, S.: Superparsing: scalable nonparametric image parsing
  with superpixels. International Journal of Computer Vision  (2012)

\bibitem{SegTrack}
Tsai, D., Flagg, M., Nakazawa, A., Rehg, J.M.: Motion coherent tracking using
  multi-label mrf optimization. International journal of computer vision
  \textbf{100}(2),  190--202 (2012)

\bibitem{ViGr09}
Vijayanarasimhan, S., Grauman, K.: What's it going to cost you?: Predicting
  effort vs. informativeness for multi-label image annotations. In: IEEE
  Conference on Computer Vision and Pattern Recognition (CVPR) (2009)

\bibitem{OnAVOS}
Voigtlaender, P., Leibe, B.: Online adaptation of convolutional neural networks
  for video object segmentation. In: British Machine Vision Conference (BMVC)
  (2017)

\bibitem{WaEtAl19}
Wang, C., Xu, D., Zhu, Y., Martin-Martin, R., Lu, C., Fei-Fei, L., Savarese,
  S.: Densefusion: 6d object pose estimation by iterative dense fusion. In: The
  IEEE Conference on Computer Vision and Pattern Recognition (CVPR) (2019)

\bibitem{WeSz17}
Wehrwein, S., Szeliski, R.: Video segmentation with background motion models.
  In: British Machine Vision Conference (BMVC) (2017)

\bibitem{LIBSVX}
Xu, C., Corso, J.J.: Libsvx: A supervoxel library and benchmark for early video
  processing. International Journal of Computer Vision  \textbf{119}(3),
  272--290 (2016)

\bibitem{YTVOS}
Xu, N., Yang, L., Fan, Y., Yue, D., Liang, Y., Yang, J., Huang, T.S.:
  Youtube-vos: {A} large-scale video object segmentation benchmark. CoRR
  \textbf{abs/1809.03327} (2018)

\bibitem{UiYamaguchi2015}
Yamaguchi, U., Saito, F., Ikeda, K., Yamamoto, T.: Hsr, human support robot as
  research and development platform. The Abstracts of the international
  conference on advanced mechatronics : toward evolutionary fusion of IT and
  mechatronics : ICAM  \textbf{2015.6},  39--40 (2015)

\bibitem{HSR_journal}
Yamamoto, T., Terada, K., Ochiai, A., Saito, F., Asahara, Y., Murase, K.:
  Development of human support robot as the research platform of a domestic
  mobile manipulator. ROBOMECH Journal  \textbf{6}(1), ~4 (2019)

\bibitem{OSMN}
Yang, L., Wang, Y., Xiong, X., Yang, J., Katsaggelos, A.K.: Efficient video
  object segmentation via network modulation. IEEE Conference on Computer
  Vision and Pattern Recognition (CVPR)  (2018)

\end{thebibliography}
\end{document}